\documentclass[manuscript]{acmart}
\AtBeginDocument{%
  }

\setcopyright{acmlicensed}
\copyrightyear{2026}
\acmYear{2026}
\acmDOI{10.1145/3787499}

\acmJournal{CSUR}
\acmVolume{58}
\acmNumber{9}
\acmArticle{225}
\acmMonth{2}

\usepackage{multirow}
\usepackage{csquotes}
\usepackage{color,colortbl}
\usepackage{pgfplots}
\usepackage{adjustbox}
\usepackage{pifont}
\usepackage{threeparttable}
\newcommand{\cmark}{\textcolor{green!60!black}{\ding{51}}}
\newcommand{\xmark}{\textcolor{red}{\ding{55}}}  
\usepackage{enumitem}
\definecolor{LightCyan}{rgb}{0.8,1,1}
\pgfplotsset{compat=1.18}
\begin{document}

\title{Large Language Models for Computer-Aided Design: A Survey}

\author{Licheng Zhang}
\authornote{Corresponding author.}
\affiliation{
  \institution{The University of Melbourne}
  \city{Parkville}
  \state{Victoria}
  \postcode{3010}
  \country{Australia}
}
\email{licheng.zhang@student.unimelb.edu.au}

\author{Bach Le}
\affiliation{
  \institution{The University of Melbourne}
  \country{Australia}
}
\email{bach.le@unimelb.edu.au}

\author{Naveed Akhtar}
\affiliation{
  \institution{The University of Melbourne}
  \country{Australia}
}
\email{naveed.akhtar1@unimelb.edu.au}

\author{Siew-Kei Lam}
\affiliation{
  \institution{Nanyang Technological University}
  \streetaddress{50 Nanyang Avenue}
  \postcode{639798}
  \country{Singapore}
}
\email{assklam@ntu.edu.sg}

\author{Tuan Ngo}
\affiliation{
  \institution{The University of Melbourne}
  \country{Australia}
}
\email{dtngo@unimelb.edu.au}

\renewcommand{\shortauthors}{Zhang et al.}

\begin{abstract}
Large Language Models (LLMs) have seen rapid advancements in recent years, with models like ChatGPT and DeepSeek, showcasing their remarkable capabilities across diverse domains. While substantial research has been conducted on LLMs in various fields, a comprehensive review focusing on their integration with Computer-Aided Design (CAD) remains notably absent. CAD is the industry standard for 3D modeling and plays a vital role in the design and development of products across different industries. As the complexity of modern designs increases, the potential for LLMs to enhance and streamline CAD workflows presents an exciting frontier. This article presents the first systematic survey exploring the intersection of LLMs and CAD. We begin by outlining the industrial significance of CAD, highlighting the need for Artificial Intelligence (AI)-driven innovation. Next, we provide a detailed overview of the foundation of LLMs. We also examine both closed-source LLMs as well as publicly available models. The core of this review focuses on the various applications of LLMs in CAD, providing a taxonomy of six key areas where these models are making considerable impact. We also provide a comprehensive study of CAD evaluation, reviewing existing methods and metrics in detail. In our analysis, we also examine common data modalities, model usage trends, dataset sources, and industrial application domains to provide a well-rounded picture of the field. Finally, we propose several promising future directions for further advancements, which offer vast opportunities for innovation and are poised to shape the future of CAD technology. Github: https://github.com/lichengzhanguom/LLMs-CAD-Survey-Taxonomy
\end{abstract}



\maketitle
\newpage
\begin{figure*}
    \centering
    \includegraphics[width=\linewidth]{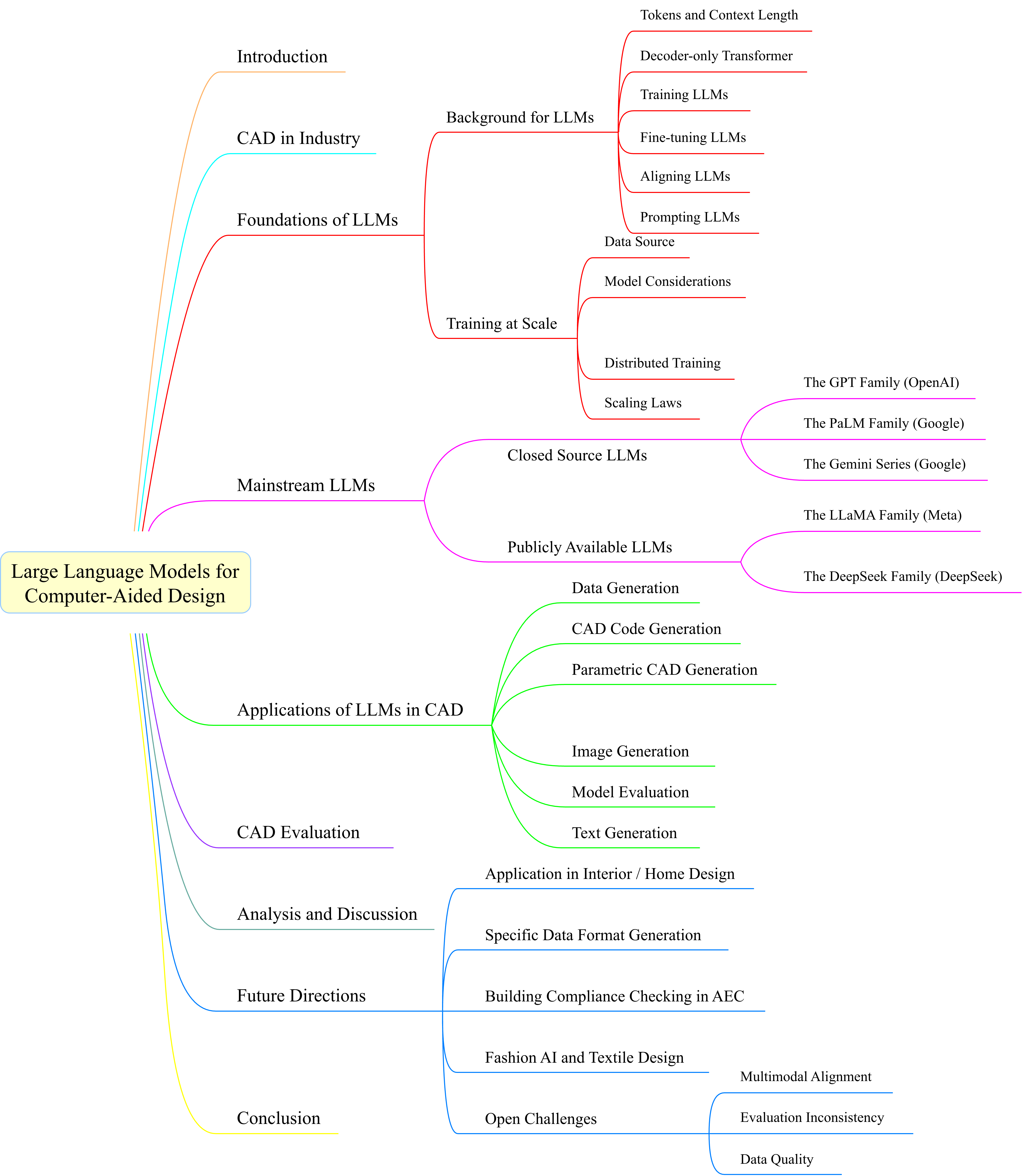}
    \caption{Taxonomy of this review.}
    \label{fig:taxonomy}
\end{figure*}
\section{Introduction}\label{sec1}
Language is a fundamental aspect of human intelligence, as essential as vision. For decades, researchers have strived to endow machines with the ability to reason and communicate in natural language \cite{zhao2023survey}. In recent years, this vision has come closer to reality with the advent of Large Language Models (LLMs), which have demonstrated remarkable progress across a wide range of tasks \cite{chang2024survey, zhao2024explainability, wanefficient, huang2023towards, wang2023aligning, das2025security, wu2024survey, zheng2025towards}. A key insight driving LLMs is that knowledge about the world can be effectively captured and represented through large-scale language modeling. This enables the construction of general-purpose models that can tackle a variety of problems via pre-training and task-specific fine-tuning, alignment, or prompting, bypassing the need for extensive domain-specific training from scratch \cite{xiao2025foundations}. LLMs are typically built upon the Transformer architecture \cite{vaswani2017attention},  with a vastly larger number of parameters compared to earlier Pre-trained Language Models (PLMs). This increase in scale has resulted in significantly improved capabilities in language understanding and generation. The release of ChatGPT \cite{chatgpt} marked a major milestone, showcasing unprecedented performance. Since then, many powerful LLMs have been introduced, including GPT-4 \cite{achiam2023gpt} and GPT-4V \cite{openai2023} by OpenAI, LLaMA \cite{touvron2023llama}, LLaMA-2 \cite{touvron2023llama2}, and LLaMA-3 \cite{grattafiori2024llama} by Meta, and GLaM \cite{du2022glam}, PaLM \cite{chowdhery2023palm}, PaLM-2 \cite{anil2023palm}, and Gemini \cite{team2023gemini} by Google, among others. Parallel to these advancements in LLMs, Computer-Aided Design (CAD) has remained a cornerstone technology in engineering and industrial design. CAD refers to the use of computers to assist in the creation, modification, analysis, or optimization of a design \cite{sarcar2008computer, meinders2008numerical, gujarathi2011parametric, quintana2010will}. It is widely used to produce 2D drawings and 3D models of physical products, supporting applications in fields such as architecture, automotive design, manufacturing, and 3D printing. CAD enhances design precision, facilitates rapid iteration, and reduces both development time and cost. Moreover, CAD systems preserve essential design information such as geometry, dimensions, and structural details in standardized file formats, promoting reusability and collaboration.

Despite the growing popularity of LLMs and the critical importance of CAD in modern industry, no survey article to date has systematically explored the intersection of these two domains. While  reviewing efforts have been made for various  LLM applications \cite{hadi2023survey, minaee2024largelanguagemodelssurvey, naveed2023comprehensive, yin2024survey, kumar2024large, annepaka2024large, hadi2024large, shen2023large, hou2024large}, a comprehensive survey  of exploiting LLMs for CAD still remains missing. On the other hand, LLMs have already demonstrated transformative potential in various generative and analytical tasks, including image synthesis \cite{brade2023promptify, koh2023generating}, text generation \cite{wu2024large, li2024pre}, and code completion \cite{cheng2025security, nejjar2025llms}. This makes them a promising candidate for advancing CAD through automation, intelligent design assistance, and semantic understanding of design specifications. Hence, a dedicated review of this emerging research area is both timely and essential. It has the potential to  offer invaluable insights for researchers and practitioners seeking to integrate LLMs into CAD workflows. This article presents the first comprehensive survey of research at the intersection of LLMs and CAD. Our objective is to consolidate recent developments, identify trends, and highlight opportunities for future research in this direction. In addition to surveying LLM applications in CAD, we also extend the scope of prior LLM surveys by incorporating the latest generation of models. Figure~\ref{fig:taxonomy} illustrates the structure and organization of this survey. Specifically, our survey makes the following key  contributions.
\begin{itemize}[leftmargin=10pt]
\item[1.] \textbf{Overview of CAD and its industrial significance:} We provide a context on the role of CAD across various industries to onboard researchers and practitioners in the LLM domain to CAD's industrial significance.
\item[2.] \textbf{Foundations of LLMs:} We review the related foundational  concepts of LLMs in an accessible manner to bring CAD experts up to speed with the fundamental knowledge necessary to appreciate the contemporary developments related to  LLMs.
\item[3.] \textbf{Review of state-of-the-art LLMs:} We provide a summary of the key recent developments in LLMs that are relevant to propel CAD developments with LLM augmentations. 

\item[4.] \textbf{Applications of LLMs in CAD:} We systematically categorize and analyze the existing literature on LLMs in CAD-related tasks.

\item[5.] \textbf{Study of CAD evaluation:} We comprehensively review existing CAD evaluation methods and metrics reported in the literature, integrating diverse approaches to assess CAD models and their practical relevance.

\item[6.] \textbf{Discussion and future directions:} We provide a detailed discussion on the limitations, challenges, and potential future pathways for integrating LLMs with  CAD. 
\end{itemize}
\section{CAD in Industry}\label{industry}
CAD involves the use of computers or workstations to assist in the creation and modification of designs, replacing traditional pencil drawings with precise digital sketches. CAD enhances designers' productivity, improves design quality, and streamlines communication through clear documentation. Notably, CAD tools can reduce design and prototyping time by 30\% to 50\%, particularly in iterative design processes \cite{camburn2017design}. CAD applications span a wide range of sectors from automotive, aerospace and defense to architecture and  manufacturing. Alin et al.~\cite{alin2024evolution} note that CAD is especially important in engineering and manufacturing, buildings and construction, medicine and biotechnology, industrial design and consumer products, and media and entertainment. Similarly, \cite{ikubanni2022present} identifies CAD applications in  electronics, shipbuilding, aerospace, the textile industry, and education. Taken together, these sources illustrate that CAD is extensively employed across a wide variety of practical, design-related domains.
Taking architectural design as a case in point, CAD is used to model buildings and infrastructure with high precision. Architects and designers can generate detailed 3D models, explore alternative design options, develop construction plans, and collaborate effectively with structural engineers and other project stakeholders \cite{alin2024evolution}. CAD also enables virtual beautification of building designs, helping to reflect realistic expectations visually \cite{ikubanni2022present}.

The global 3D CAD software market reflects this widespread adoption. This market is projected to grow from USD 13.40 billion in 2025 to USD 24.23 billion by 2034\footnote{\url{https://www.towardspackaging.com/insights/3d-cad-software-market-sizing?utm_source=chatgpt.com}}. This growth is largely driven by technological advancements. In 2023, North America led the global CAD software market, while the Asia-Pacific region was forecasted to experience significant growth through 2034. It is noteworthy that the on-premises deployment model accounted for the largest market share in 2023. The large enterprise segment is anticipated to expand significantly between 2025 and 2034. Moreover, the AEC (Architecture, Engineering, and Construction) segment dominated the CAD software market in 2023. These facts present a clear picture of industrial significance of CAD based design.
\section{Foundations of LLMs}\label{llms}
In this section, we discuss foundational concepts related to LLMs in an accessible manner. Our aim is to introduce the relevant terms and concepts to non-LLM experts to bridge the gap between LLM and CAD research communities. This section is not aimed at discussing the details of these concepts. Readers interested in technical details are referred to \cite{naveed2023comprehensive, zhao2023survey}.
\subsection{Background for LLMs}
\subsubsection{Tokens and Context Length}
In the context of LLMs, tokens are the basic units of text that the model reads, processes, and generates. Tokens are pieces of text, like words, subwords, word pieces, or characters depending on the tokenizer. The context window, also known as context length, refers to the maximum number of tokens an LLM can see or process at once when generating or understanding text.
\subsubsection{Decoder-only Transformer}
The decoder-only Transformer architecture \cite{vaswani2017attention} is one of the most widely used structures for building LLMs. Its core structure consists of a stack of Transformer blocks \cite{vaswani2017attention}, where each block includes two sub-layers: one for self-attention modeling and another for feed-forward network modeling. A softmax layer is placed on top of the final Transformer block to generate a probability distribution over the vocabulary. Self-attention, feed-forward networks (FFNs) and softmax layer are all standard components of contemporary neural networks.  Although implementation details may vary, many LLMs share this general  architecture. These models are referred to as ``large'' due to their significant width and depth \cite{xiao2025foundations}.
\subsubsection{Training LLMs}
Training LLMs involves the standard neural network optimization process, typically using gradient descent algorithms \cite{krizhevsky2012imagenet}. Studies have shown that model performance improves as they are trained on larger datasets, and when they are scaled up in terms of architecture and computational capacity \cite{kaplan2020scaling}. This insight has led to continued efforts to increase both the size of training data and model complexity, resulting in increasingly powerful contemporary LLMs.
\subsubsection{Fine-tuning LLMs}
LLMs are first pre-trained on a large corpus of textual data. This pre-training achieves general-purpose language understanding and generation capabilities. Commonly, this is followed by fine-tuning these models to solve specific natural language processing (NLP) tasks. This fine-tuning involves further training of the model on a limited data, which is available for the downstream task. However, fine-tuning only changes the model parameters slightly, thereby making the process of adaptation computationally less expensive. Since the fine-tuning process may also use instruction-following data, such fine-tuning is also known as instruction fine-tuning. To enable instruction-following, datasets containing various instructions and corresponding responses are required. Scaling the number of such tasks for fine-tuning generally improves model performance \cite{chung2024scaling}. Unlike pre-training, which may require billions or trillions of samples, fine-tuning can be performed with tens or hundreds of thousands of high-quality samples \cite{chenalpagasus}. Fine-tuning plays a central role in enabling and enhancing this versatility, and ongoing research continues to improve fine-tuning techniques to make LLMs more efficient and effective.
\subsubsection{Aligning LLMs}
Alignment refers to the process of guiding LLMs to behave in accordance with human intentions and ethical standards. This often involves incorporating human feedback, labeled data, or explicitly defined preferences into the training process. Alignment is essential to ensure responsible and safe artificial intelligence (AI) behavior. Typically, alignment follows two main steps after initial pre-training:
\begin{itemize}[leftmargin=10pt]
    \item[1.] Supervised Fine-tuning (SFT): This step usually uses instruction-based data to further refine the model.
    \item[2.] Reinforcement Learning from Human Feedback (RLHF) \cite{ouyang2022training}: In this phase, alignment is treated as a reinforcement learning (RL) problem \cite{shakya2023reinforcement}. A reward model representing the environment evaluates outputs based on human feedback, while the LLM acts as the agent being optimized to maximize these rewards.
\end{itemize}

These techniques are critical to adapting LLMs for real-world applications, especially where safety, ethical behavior, and user alignment are essential.
\subsubsection{Prompting LLMs}
Prompting plays a key role in effective utilization of LLMs, as it requires no additional training or fine-tuning. LLMs are highly versatile once they are pre-trained on large-scale datasets, and careful prompting can strongly exploit this versatility. Instead of building task-specific systems, users can simply provide well-crafted prompts to perform a wide range of tasks. Consequently, prompt engineering \cite{perez2021true} has become an active area of research within the NLP community. Owing to the benefits of prompting, LLMs are also known for their zero-shot learning ability to perform tasks they were not explicitly trained on. Another relevant concept in this regard is in-context learning (ICL). In ICL, a well-trained LLM is provided an example of the input-output mapping in the prompt itself during the model inference stage. The model tries to understand the semantics behind this mapping and generalize it to the query in the prompt. The benefit of ICL is that it does not require further training or fine-tuning of the model on the new input-output mapping. As well-known, LLMs have still been reported to face challenges in tasks that require arithmetic reasoning \cite{yang2024arithmetic} and commonsense reasoning \cite{ismayilzada2023crow}. 
This shortcoming can be helped with ICL. Nevertheless, the `reasoning' requirements of such tasks call for more sophisticated solutions.  
Hence, to incorporate reasoning capabilities, researchers use Chain-of-Thought (CoT) prompting, which encourages models to solve problems by breaking them down into a series of intermediate reasoning steps. This mimics human-like cognitive processes and has shown notable improvements, particularly in complex mathematical reasoning tasks \cite{wang-etal-2024-boosting-language}. There are three common forms of CoT prompting used in ICL:
\begin{itemize}[leftmargin=10pt]
    \item Zero-shot CoT prompting, where the model is prompted without examples but encouraged to generate intermediate reasoning steps.
    \item One-shot CoT prompting, where one example is provided in the prompt.
    \item Few-shot CoT prompting, where a small number of examples are included.
\end{itemize}

These prompting strategies enable LLMs to generalize better and solve more complex problems without modifying the model's internal parameters.
\subsection{Training at Scale}
\subsubsection{Data Source}
Data is a foundational element in training LLMs, and its importance in learning effective models cannot be overstated. While increasing the quantity of training data is essential, more data does not necessarily equate to better model performance. Several challenges must be addressed when sourcing the data:
\begin{itemize}[leftmargin=10pt]
    \item Data Quality: Raw data collected from diverse sources is often noisy or irrelevant. Therefore, filtering and cleaning processes are critical during data preparation. As demonstrated in \cite{penedo2023refinedweb}, up to 90\% of web-scraped data may need to be removed before training the model due to its adverse effect on the model.
    \item Data Diversity: A robust dataset must encompass a wide variety of domains and languages to ensure generalization and reduce bias. Data sourced from a  single or similar corpus can compromise the versatility of the model.  
    \item Bias: Bias can arise from class imbalance or insufficient representation across languages and topics in the dataset. To mitigate bias, datasets are often needed to be explicitly  balanced and diversified.
    \item Privacy and Copyright Concerns: When utilizing large-scale datasets, protecting sensitive information and respecting copyright are key considerations. LLMs can be fine-tuned to detect and refuse prompts that might lead to data leakage \cite{wu2023depn} or copyright violations \cite{karamolegkou2023copyright}.
\end{itemize}

Pre-training corpora to train LLMs at scale are typically divided into two categories: general data and specialized data \cite{zhao2023survey, hadi2023survey}. We summarize the common sources of these categories below.
\begin{itemize}[leftmargin=10pt]
    \item \textbf{General Data:}
    \begin{itemize}[leftmargin=1pt]
        \item[1.] Webpages: The Internet offers a massive corpus of both high- and low-quality text. However, filtering is vital to remove spam, misinformation, or irrelevant content while preserving valuable sources.
        \item[2.] Conversational Text: Public dialogue datasets \cite{baumgartner2020pushshift, roller2021recipes} and social media platforms provide conversational data useful for training models in dialogue and response generation.
        \item[3.] Books: Long-form text from books supports the learning of linguistic richness, narrative flow, and long-term dependencies. Datasets like Books3 and BookCorpus2 (from the Pile \cite{gao2020pile}) are commonly used.
    \end{itemize}
    \item \textbf{Specialized Data:}
    \begin{itemize}[leftmargin=1pt]
        \item[1.] Multilingual Text: They enhance the multilingual capabilities of LLMs.
        \item[2.] Scientific Texts: They strengthen domain knowledge in technical or academic contexts, often sourced from arXiv, scientific textbooks, or math websites.
        \item[3.] Code: Programming data improves the model's ability to generate structured code and solve logic-based tasks. Common sources include Stack Exchange \cite{xu2022systematic} and GitHub. Interestingly, formulating reasoning problems as code can also improve the accuracy of generated responses \cite{madaan2022language}.
    \end{itemize}
\end{itemize}
\subsubsection{Model Considerations}
Developing LLMs at scale often involves careful considerations related to the underlying neural architecture of the models. Key considerations in this regard include:
\begin{itemize}[leftmargin=10pt]
    \item Layer normalization and residual connections \cite{xiong2020layer}: Most LLMs need to apply layer normalization inside residual blocks to enhance stability and trainability of their deep Transformer-based architectures, referred as pre-layer normalization (Pre-LN). Post-layer normalization (Post-LN) can achieve better performance in shallower models, but it may lead to training instability in deeper architectures due to issues like vanishing gradients \cite{xiong2020layer}. Recent research has explored hybrid strategies to combine the benefits of both Pre-LN and Post-LN \cite{kim2025peri}.
    \item Activation functions: Activation functions are responsible for incorporating non-linearity in the modeling process of neural networks. The choice of activation function is crucial, particularly for FFNs. While ReLU \cite{glorot2011deep} is the prevailing standard, other options have also  proven effective in the context of LLMs, e.g.,  Gaussian Error Linear Unit (GeLU) \cite{hendrycks2016gaussian}, used in GPT-3 \cite{brown2020language} and BLOOM \cite{le2023bloom}, Gated Linear Unit (GLU) \cite{dauphin2017language}, used in Gemma \cite{team2024gemma}, and SwiGLU \cite{shazeer2020glu}, adopted in models like PaLM \cite{chowdhery2023palm} and LLaMA \cite{touvron2023llama}.
    \item Bias removal: Inclusion/removal of bias terms is another relevant design choice for LLM architectures.  Some models, such as LLaMA \cite{touvron2023llama} and Gemma \cite{team2024gemma}, eliminate bias terms in components like layer normalization, FFNs and query-key-value transformations to improve training dynamics.
\end{itemize}
\subsubsection{Distributed Training}
Training LLMs at scale demands significant computational resources, typically provided by distributed systems. To that end, various forms of parallelism are employed, including data parallelism, model parallelism, tensor parallelism and pipeline parallelism. For instance, the 405 billion (B) parameter version of LLaMA-3 \cite{grattafiori2024llama} was trained on up to 16000 H100 GPUs, with each server containing 8 GPUs. Communication across servers was managed using distributed parallelism.
\subsubsection{Scaling Laws}
Scaling laws \cite{kaplan2020scaling} describe the empirical relationship between LLM performance and key training variables such as model size, dataset size, and computational budget. Research shows that performance consistently improves with increased data, even at the scale of trillions of tokens, and larger models.
These scaling laws serve as a blueprint for LLM development, helping researchers make informed decisions about how to allocate resources effectively.

In summary, LLMs are advanced AI systems that can simulate human-like intelligence through large-scale learning \cite{dergaa2023human}. By leveraging deep learning architectures and massive corpora of data, they learn complex patterns in text, enabling them to generate coherent, contextually appropriate responses and content \cite{hoffmann2022training, li2022conditional}.
\section{Mainstream LLMs}\label{classic}
LLMs are a type of Transformer-based PLMs with tens to hundreds of billions of parameters. In this section, we categorize LLMs into closed-source and publicly available models. Within each category, we further group models by their families and present them according to their development timeline.
\subsection{Closed Source LLMs}
\subsubsection{The GPT Family (OpenAI)}
The Generative Pre-trained Transformer (GPT) family consists of decoder-only Transformer models developed by OpenAI. This family of LMMs started with GPT-1 \cite{radford2018improving} and GPT-2 \cite{radford2019language}, and moved on to more advanced models, discussed below. 

\textit{\textbf{GPT-3 \cite{brown2020language}:}} Often regarded as the first true LLM, GPT-3 is an autoregressive model with 175 billion parameters. It shares the same architecture as GPT-2 but introduces sparse attention and adopts gradient noise scaling \cite{mccandlish2018empirical} for better training. GPT-3 is significantly larger than earlier PLMs and exhibits emergent capabilities not seen in smaller models. \textit{\textbf{CODEX \cite{chen2021evaluating}:}} A descendant of GPT-3, Codex is fine-tuned on code corpora from GitHub and is optimized for generating code from natural language prompts. \textit{\textbf{WebGPT \cite{nakano2021webgpt}:}} Another GPT-3 variant, WebGPT is trained to answer open-ended questions by browsing web pages, enhancing its ability to perform web-based information retrieval. \textit{\textbf{InstructGPT \cite{ouyang2022training}:}} This version aligns the model with human intent by fine-tuning it using datasets containing human-written demonstrations and preferences. It shows improved truthfulness and reduced toxicity. \textit{\textbf{GPT-3.5 \cite{openai_gpt35}:}} Built on GPT-3 and Codex, GPT-3.5 benefits from fine-tuning on code and improved instruction-following as well as RLHF. GPT-3.5 Turbo \cite{openai2023gpt35turbo} is a faster, more cost-effective version optimized for chat. \textit{\textbf{ChatGPT \cite{chatgpt}:}} A major milestone in LLMs, ChatGPT was fine-tuned with human preference data sourced from a wide range of texts (e.g., Wikipedia, books, websites, scientific papers, articles and news media \cite{welsby2023chatgpt}). It excels at delivering engaging, natural conversations \cite{waghmare2023introduction} and completing diverse tasks such as translation, summarization, and Q\&A \cite{shafik2024introduction}. \textit{\textbf{GPT-4 \cite{achiam2023gpt}:}} GPT-4 is a multimodal model capable of processing both text and images. It delivers significantly improved performance on complex tasks and achieves human-level results on professional benchmarks. GPT-4 uses \enquote{predictable scaling} to optimize training efficiency, which improves performance in a measurable manner with respect to the required  training resources. GPT-4 Turbo \cite{openai2023gpt4turbo} is an enhanced version of GPT-4 with better capabilities, larger context length, and integrated tools such as vision, DALL$\cdot$E 3, and text-to-speech. \textit{\textbf{GPT-4V \cite{openai2023}:}} GPT-4V focuses on safely deploying the visual capabilities of GPT-4, offering strong performance in a variety of vision tasks. \textit{\textbf{GPT-4o \cite{hurst2024gpt}:}} GPT-4o is an \enquote{omni} model that handles any combination of text, audio, image, and video as input and output. It brings powerful multimodal capabilities with an emphasis on audio. \textit{\textbf{OpenAI o1 and o3-mini \cite{jaech2024openai, o3mini}:}} o1 uses a novel optimization technique and a tailored dataset, supporting visual reasoning, while o3-mini is a lightweight model optimized for reasoning and cost-efficiency, though it does not support vision. \textit{\textbf{GPT-4.1 \cite{openai2025gpt4.1}:}} GPT-4.1 provides near-frontier reasoning with greatly improved speed, efficiency, and reliability, making it ideal for scalable real-world applications. \textit{\textbf{GPT-4.5 \cite{openai2025gpt45}:}} Building on the advancements of GPT-4o, GPT-4.5 offers a more natural interaction experience. It features a broader knowledge base, a stronger alignment with user intent, and improved emotional intelligence, allowing it to better understand and respond to emotional cues. Additionally, GPT-4.5 reduces the occurrence of hallucinations, ensuring more accurate and reliable responses. \textit{\textbf{GPT-5 / 5.2 \cite{openai2025gpt5}:}} GPT-5 delivers cutting-edge reasoning, advanced multimodal understanding, and strong long-context performance, enabling highly accurate and reliable real-world AI autonomy. GPT-5.2 features improved general reasoning, stronger multimodal understanding, and enhanced long-context performance compared with earlier versions.
\subsubsection{The PaLM Family (Google)}
The PaLM (Pathways Language Model) series is developed by Google AI.

\textit{\textbf{PaLM \cite{chowdhery2023palm}:}} It is a 540B-parameter model trained on 780B tokens. It introduces several architectural enhancements like SwiGLU activation, multi-query attention, and shared input-output embeddings. PaLM shows state-of-the-art few-shot performance on a wide range of tasks. \textit{\textbf{PaLM-2 \cite{anil2023palm}:}} PaLM-2 is a smaller (340B-parameter) but more efficient model, pre-trained on 3.6 trillion (T) tokens. It delivers superior reasoning and multilingual capabilities while reducing training and inference costs. \textit{\textbf{Med-PaLM / Med-PaLM 2 \cite{singhal2023large, singhal2025toward}:}} Med-PaLM is a domain-specific model fine-tuned for medical tasks. Med-PaLM 2 outperforms its predecessor using domain adaptation and ensemble techniques, achieving strong performance across clinical benchmarks and professional exams. \textit{\textbf{PaLM-E \cite{driess2023palm}:}} It is a multimodal version with 562B parameters (540B from PaLM and 22B from a vision Transformer \cite{dehghani2023scaling}). It supports visual question answering, zero-shot multimodal CoT reasoning, few-shot prompting, OCR-free math reasoning, and multi-image reasoning. \textit{\textbf{U-PaLM \cite{tay2023transcending}:}} U-PaLM improves model quality while requiring minimal additional computational resources, achieving similar or better performance than PaLM at about half the computational budget. \textit{\textbf{Flan-PaLM / Flan-U-PaLM \cite{chung2024scaling}:}} These versions focus on instruction tuning and CoT prompting. Flan-PaLM, with 540B parameters fine-tuned on 1.8K tasks, significantly outperforms the base PaLM model. Flan-U-PaLM further enhances the performance with advanced adaptation techniques like UL2R \cite{tay2023transcending}.
\subsubsection{The Gemini Series (Google)}
Gemini is a new family of multimodal LLMs, succeeding LaMDA \cite{thoppilan2022lamda} and PaLM-2.

\textit{\textbf{Gemini \cite{team2023gemini}:}} Gemini demonstrates strong capabilities across text, image, video, and audio understanding, with support for a 32K context length. \textit{\textbf{Gemini 1.5 \cite{team2024gemini}:}} Gemini 1.5 brings notable advancements, including a new model architecture, the use of mixture-of-experts (MoE), and training on a much larger dataset containing millions of tokens. \textit{\textbf{Gemini 2.0 \cite{pichai2024introducing}:}} Gemini 2.0 is a major update to its predecessor, improving both speed and performance compared to Gemini 1.5. \textit{\textbf{Gemini 2.5 \cite{gemini25}:}} Gemini 2.5 further enhances reasoning and coding capabilities, incorporating techniques like CoT prompting for more structured reasoning. \textit{\textbf{Gemini 3 \cite{chatlyai2025gemini3pro}:}} Gemini 3 is Google DeepMind’s AI model offering further advanced reasoning, creative problem-solving, and reliable multimodal understanding. However, Gemma \cite{team2024gemma} and Gemma 2 \cite{team2024gemma2} are lightweight, open-source versions of Gemini, providing powerful multimodal capabilities for free.

We summarize state-of-the-art closed-source LLMs in Table \ref{tab:closellms}. Since ChatGPT, released in November 2022, was the first LLM to demonstrate exceptional performance across a variety of tasks, we focus on models released from 2023 onward. Before ChatGPT, LLMs were generally less powerful, but following its release, models have grown powerful enough to compete with ChatGPT, driven by exponential growth in both model size and data.
It is worth noticing that we only list representative LLMs in the table. For example, we list Gemini but omit Gemini 1.5 due to limited technical innovations.  However, since PaLM and PaLM-2 represent distinct models, both are included.

\begin{table}[t]
\centering
\caption{Summary of state-of-the-art closed-source LLMs.}
\label{tab:closellms}
\renewcommand{\arraystretch}{1}
\begin{tabular}{lccccl}
\hline
\rowcolor{LightCyan}
Model & \#Param. & Tokens & Year & Entity & Key Feature \\
\hline
\hline
ChatGPT \cite{chatgpt} & - & - & 2022 & OpenAI & Dialogue-optimized LLM via RLHF \\
GPT-4 \cite{achiam2023gpt} & - & - & 2023 & OpenAI & Multimodal support, stronger reasoning \\
GPT-4V \cite{openai2023} & - & - & 2023 & OpenAI & Vision-enabled GPT-4 variant \\
GPT-4o \cite{hurst2024gpt} & 200B & - & 2024 & OpenAI & Omni-modal (text, image, audio, video) \\
OpenAI o3-mini \cite{o3mini} & 3B & - & 2025 & OpenAI & Lightweight GPT-4 variant for mobile use \\
GPT-4.5 \cite{openai2025gpt45} & - & - & 2025 & OpenAI & Improved emotional intelligence \\
PaLM \cite{chowdhery2023palm} & 540B & 780B & 2023 & Google & Dense decoder-only model \\
PaLM-2 \cite{anil2023palm} & 340B & 3.6T & 2023 & Google & Fine-tuned for multilingual reasoning \\
Med-PaLM \cite{singhal2023large} & 540B & - & 2023 & Google & MedQA \cite{jin2021disease} benchmarked medical LLM \\
PaLM-E \cite{driess2023palm} & 562B & - & 2023 & Google & Embodied multimodal reasoning \\
U-PaLM \cite{tay2023transcending} & 540B & - & 2023 & Google & Unified fine-tuning with few-shot and CoT \\
Flan-PaLM \cite{chung2024scaling} & 540B & - & 2024 & Google & Instruction-tuned PaLM variant \\
Med-PaLM 2 \cite{singhal2025toward} & - & - & 2025 & Google & Safety-aligned medical LLM \\
Gemini \cite{team2023gemini}& - & - & 2023 & DeepMind & Multimodal, integrated with search \\
PanGu-$\Sigma$ \cite{ren2023pangu} & 1085B & 329B & 2023 & Huawei & China's largest LLM, multilingual \\
BloombergGPT \cite{wu2023bloomberggpt} & 50B & 708B & 2023 & Bloomberg & Financial-domain LLM \\
PPLX & 70B & - & 2023 & Perplexity AI & Retrieval-augmented, mobile-ready LLM \\
Inflection-2.5 & 2.5B & - & 2024 & Inflection AI & Personal assistant, dialogue-tuned \\
Claude 3/3.5/4/4.5 \cite{TheC3} & - & - & 2024 & Anthropic & Constitutional AI, safety-tuned \\
Grok 3\&4 & - & - & 2025 & xAI & Twitter-integrated, humor-oriented chat \\
\hline
\end{tabular}
\end{table}
\subsection{Publicly Available LLMs}
\subsubsection{The LLaMA Family (Meta)}
The LLaMA family consists of a series of foundational language models developed by Meta.

\textit{\textbf{LLaMA \cite{touvron2023llama}:}} The first model in the LLaMA family, LLaMA, ranges from 7B to 65B parameters and is pre-trained on trillions of tokens. It is built upon the Transformer architecture, similar to GPT-3 [18], with several architectural modifications, including:
\begin{itemize}[leftmargin=10pt]
    \item Adoption of the SwiGLU activation function instead of ReLU.
    \item Replacement of absolute positional embeddings with rotary positional embeddings.
    \item Use of root mean squared layer normalization rather than standard layer normalization.
\end{itemize}

\textit{\textbf{LLaMA-2 \cite{touvron2023llama2}:}} LLaMA-2 includes both foundational models and chat-optimized versions, such as LLaMA-2 Chat. While the architecture remains largely the same as LLaMA, LLaMA-2 was trained on 40\% more data. LLaMA-2 Chat further improves safety by fine-tuning on safe response samples and incorporating an additional RLHF step. \textit{\textbf{LLaMA-3 / 3.1 \cite{grattafiori2024llama}:}} LLaMA-3 / 3.1 models are trained on a dataset seven times larger than that of LLaMA-2. LLaMA-3 comes in two sizes: 8B and 70B. With the largest model featuring 405B parameters, LLaMA-3.1 benefits from 15T tokens of training data and a significantly larger context window (128K compared to LLaMA-2’s 8K). These improvements lead to its competitive performance, achieving results comparable to GPT-4 and GPT-4o. \textit{\textbf{LLaMA-4 \cite{meta2025llama4}:}} LLaMA-4 introduces significant advancements by supporting text, images, audio, and video, enabling seamless multimodal reasoning and generation. This is a major shift from its predecessors, LLaMA-2 and LLaMA-3, which were primarily text-based. Leveraging a Mixture of Experts (MoE) architecture, LLaMA-4 efficiently scales to trillions of parameters, activating only a subset of experts during inference for optimal performance. Additionally, it supports an impressive context length of up to 10 million tokens, making it ideal for long-form tasks like multi-document summarization and complex dialogues.

In addition to the core LLaMA models, a variety of instruction-following models have been developed based on LLaMA or LLaMA-2, including: Alpaca \cite{taori2023alpaca}, Vicuna \cite{chiang2023vicuna}, Guanaco \cite{dettmers2023qlora}, Koala \cite{koala}, Mistral 7B \cite{jiang2023mistral7b}, Code LLaMA \cite{roziere2023code}, Gorilla \cite{patil2024gorilla}, Giraffe \cite{pal2023giraffe}, Vigogne \cite{vigogne}, Tulu 65B \cite{wang2023far}, Long LLaMA \cite{tworkowski2023focused}, Stable Beluga 2 \cite{StableBelugaModels}, Qwen \cite{bai2023qwen}, among others.
Vicuna is particularly popular for its multimodal language modeling, giving rise to models like LLaVA \cite{liu2023visual}, MiniGPT-4 \cite{zhu2024minigpt}, InstructBLIP \cite{dai2023instructblipgeneralpurposevisionlanguagemodels}, and PandaGPT \cite{su2023pandagpt}.
\begin{table}[t]
\footnotesize
\centering
\caption{Summary of state-of-the-art publicly available LLMs.}
\label{tab:publlms}
\renewcommand{\arraystretch}{1}
\begin{tabular}{lccccl}
\hline
\rowcolor{LightCyan}
Model & \#Params. & Tokens & Year & Entity & Key Feature \\
\hline
\hline
LLaMA \cite{touvron2023llama} & 65B & 1.4T & 2023 & Meta & Foundation model for instruction tuning \\
LLaMA-2 \cite{touvron2023llama2} & 70B & 2T & 2023 & Meta & Improved safety, performance, training data \\
Code LLaMA \cite{roziere2023code} & 70B & 1T & 2023 & Meta & Code-specific extension of LLaMA \\
LIMA \cite{zhou2023lima} & 65B & - & 2023 & Meta & Fine-tuned with only 1K examples\\
Long LLaMA \cite{tworkowski2023focused} & 7B & 1T & 2023 & Meta & Long-context with Focused Transformer \cite{tworkowski2023focused} \\
LLaMA-3 \cite{grattafiori2024llama} & 70B & 15T & 2024 & Meta & Supporting longer context, enhanced reasoning \\
LLaMA-4 \cite{meta2025llama4} & 109B & 40T & 2025 & Meta & 10 million context length, multimodal $+$ MoE \\
DeepSeek-Coder \cite{guo2024deepseek} & 33B & 2T & 2024 & DeepSeek & Code synthesis and reasoning \\
DeepSeek-LLM \cite{bi2024deepseek} & 67B & 2T & 2024 & DeepSeek & General-purpose pre-trained \\
DeepSeek-MoE \cite{dai2024deepseekmoe} & 16B & 2T & 2024 & DeepSeek & Sparse MoE \\
DeepSeek-Math \cite{shao2024deepseekmath} & 7B & 120B & 2024 & DeepSeek & Math benchmark specialist \\
DeepSeek-VL \cite{lu2024deepseek} & 7B & 2T & 2024 & DeepSeek & Vision-language aligned \\
DeepSeek-V2 \cite{liu2024deepseek} & 236B & 8.1T & 2024 & DeepSeek & High-performance general model \\
DeepSeek-V3/3.1 \cite{liu2024deepseekv3} & 671B & 14.8T & 2024 & DeepSeek & Flagship MoE LLM \\
DeepSeek-R1 \cite{guo2025deepseek} & 70B & - & 2025 & DeepSeek & Reinforced instruction tuning \\
Stable Beluga 2 \cite{StableBelugaModels} & 70B & - & 2023 & Stability AI & Instruction-tuned LLaMA-2 \\
Stable LM 2 \cite{bellagente2024stable} & 1.6B & 2T & 2024 & Stability AI & Lightweight model \\
Giraffe \cite{pal2023giraffe} & 13B & - & 2023 & Salesforce & Visual grounding and reasoning \\
InstructBLIP \cite{dai2023instructblipgeneralpurposevisionlanguagemodels} & 13B & - & 2023 & Salesforce & General-purpose vision-language model \\
CodeGen2 \cite{nijkamp2023codegen2} & 16B & 400B & 2023 & Salesforce & Multi-language code generation \\
Koala \cite{koala} & 13B & - & 2023 & Berkeley & Dialogue-tuned LLaMA on curated data \\
Gorilla \cite{patil2024gorilla} & 7B & - & 2024 & Berkeley & API call-focused LLM \\
Orca \cite{mukherjee2023orca} & 13B & - & 2023 & Microsoft & Imitation learning from GPT-4 \\
WizardLM \cite{xu2024wizardlm} & 13B & - & 2024 & Microsoft & Evol-Instruct fine-tuned \\
Alpaca \cite{taori2023alpaca} & 7B & - & 2023 & Stanford & Instruction-tuned LLaMA on self-instruct \\
Vicuna \cite{chiang2023vicuna} & 13B & - & 2023 & LMSYS & Chatbot fine-tuned on ShareGPT \cite{sharegptvicuna2023dataset} \\
Guanaco \cite{dettmers2023qlora} & 65B & - & 2023 & UW & QLoRA \cite{dettmers2023qlora} fine-tuned Vicuna \\
Mistral \cite{jiang2023mistral7b} & 7B & - & 2023 & Mistral AI & Sliding window attention, strong performance \\
Vigogne \cite{vigogne} & 7B & - & 2023 & - & French fine-tuned Vicuna \\
Tulu \cite{wang2023far} & 65B & 1.4T & 2023 & Allen AI & Instruction-tuned LLaMA \\
Qwen \cite{bai2023qwen} & 14B & 3T & 2023 & Alibaba & Multilingual and code generation \\
LLaVA \cite{liu2023visual} & 13B & - & 2023 & UIUC & Vision-language instruction tuning \\
PandaGPT \cite{su2023pandagpt} & 13B & - & 2023 & - & Multimodal instruction-following \\
Pythia \cite{biderman2023pythia} & 12B & 300B & 2023 & EleutherAI & Transparent training for scientific benchmark \\
Baichuan2 \cite{yang2023baichuan} & 13B & 2.6T & 2023 & Baichuan & Strong multilingual and reasoning performance \\
FLM \cite{li2023flm} & 101B & 311B & 2023 & - & Hybrid MoE $+$ dense \\
Skywork \cite{wei2023skywork} & 13B & 3.2T & 2023 & - & Multilingual, instruction-tuned \\
Falcon \cite{penedo2023refinedweb} & 7.5B & 600B & 2023 & TII & RefinedWeb dataset \\
Zephyr \cite{tunstallzephyr} & 7.24B & 800B & 2023 & HuggingFace & Aligned LLaMA-2 variant \\
StartCoder \cite{li2023starcoder} & 15.5 & 1T & 2023 & - & Code generation LLM \\
MPT \cite{MosaicML2023Introducing} & 7B & 1T & 2023 & MosaicML & Commercially permissive LLM \\
XuanYuan 2.0 \cite{zhang2023xuanyuan} & 176B & - & 2023 & - & Financial sector \\
CodeT5$+$ \cite{wang-etal-2023-codet5} & 16B & 51.5B & 2023 & - & Code understanding $+$ generation \\
BloomZ and mT0 \cite{muennighoff2023crosslingual} & 176B & 350B & 2023 & BigScience & Cross-lingual capabilities \\
Jamba \cite{lieber2024jamba} & 52B & - & 2024 & AI21 Labs & Hybrid Transformer $+$ MoE \\
Gemma \cite{team2024gemma} & 7B & 6T & 2024 & Google & Lightweight model with strong benchmarks \\
MiniGPT-4 \cite{zhu2024minigpt} & 13B & - & 2024 & - & Image-to-text multimodal capabilities \\
ChatGLM \cite{glm2024chatglm} & 9B & 10T & 2024 & Tsinghua & Bilingual chat, long-context \\
Llemma \cite{azerbayev2024llemma} & 34B & 50B & 2024 & - & Math and science focused \\
Command R$+$\tnote{d} & 104B & - & 2024 & Cohere & RAG-enhanced instruction model for finance \\
\hline
\end{tabular}
\end{table}
\subsubsection{The DeepSeek Family (DeepSeek)}
The DeepSeek series consist of LLMs developed by the Chinese AI company DeepSeek.

\textit{\textbf{DeepSeek-Coder \cite{guo2024deepseek}:}} DeepSeek-Coder is the first model in the DeepSeek family, followed by DeepSeek-LLM \cite{bi2024deepseek}, DeepSeek-MoE \cite{dai2024deepseekmoe}, and DeepSeek-Math \cite{shao2024deepseekmath}. DeepSeek-LLM investigates the scaling laws for LLMs to identify the optimal model size and training data scale. DeepSeek-MoE is an innovative MoE architecture specially designed towards ultimate expert specialization. DeepSeek-Math is a domain-specific language model that significantly outperforms the mathematical capabilities of open-source models and approaches the performance level of GPT-4 on academic benchmarks. \textit{\textbf{DeepSeek-VL \cite{lu2024deepseek}:}} DeepSeek-VL is an open-source Vision-Language Model designed for real-world applications that combine vision and language understanding. \textit{\textbf{DeepSeek-V2 \cite{liu2024deepseek}:}} DeepSeek-V2 uses multi-head latent attention to reduce inference costs by compressing the key-value cache into a latent vector. This method achieves 5.76 times faster inference throughput compared to the previous DeepSeek models. \textit{\textbf{DeepSeek-V3 / 3.1 \cite{liu2024deepseekv3}:}} DeepSeek-V3 is a stronger MoE model, featuring 671B parameters, with 37B activated per token. It is pre-trained on 14.8T high-quality, diverse tokens and undergoes SFT and RL stages to fully optimize its capabilities. Compared to DeepSeek-V3, DeepSeek-V3.1 doubles the context window, introduces a hybrid reasoning mode, enhances math, logic, and code capabilities, and improves multimodal performance while maintaining efficient, open deployment. \textit{\textbf{DeepSeek-R1 \cite{guo2025deepseek}:}} DeepSeek-R1-Zero is pre-trained with large-scale RL without the initial SFT stage. DeepSeek-R1 follows a multi-stage training process, including cold-start data before RL, and incorporating two RL stages to improve reasoning patterns and align with human preferences, along with two SFT stages to enhance reasoning and non-reasoning abilities.

We present an overview of state-of-the-art publicly available LLMs in Table \ref{tab:publlms},
focusing on listing only the most representative LLMs of each type.
\section{Applications of LLMs in CAD}\label{application}
This section presents the existing literature focusing on applications of LLMs in CAD. Due to the nascency of this emerging but critical   research area, we are able to provide a comprehensive overview of the recent developments. These advances are organized by grouping the works according to their key application directions  in the CAD domain.
\subsection{Data Generation}
LLMs have emerged as powerful tools for content generation, making data generation one of their primary applications in the field of CAD. This capability is particularly useful for creating synthetic datasets, fine-tuning models, and addressing the lack of annotated CAD data. Yuan et al. \cite{YUAN2024104048} utilized GPT-4o \cite{hurst2024gpt} to generate datasets that map natural language descriptions and rendered view images to CAD operation sequences. This approach enabled the construction of textual descriptions corresponding to CAD commands. Similarly, Xu et al. \cite{xu2025cadmllmunifyingmultimodalityconditionedcad} employed the open-source model InternVL2-26B \cite{chen2024internvl}, by randomly selecting four view images of CAD models to generate high-quality textual captions. These captions were used to build a multimodal CAD dataset. Khan et al. \cite{khan2024text2cad} adopted a two-step generation process using open-source LLMs and Vision-Language Models (VLMs). First, they rendered multi-view images of individual parts and the final models, which were input to  LLaVA-NeXT model \cite{liu2024llavanext} with predefined prompts to generate simplified object-level descriptions. Then, they used Mixtral-50B \cite{jiang2024mixtral} to produce multi-level natural language instructions. Raw CAD sequences from DeepCAD \cite{wu2021deepcad} were preprocessed to replace meaningless keys with more descriptive terms, which, along with the simplified shape descriptions, served as input to Mixtral-50B. The result was a four-level hierarchy of textual instructions: abstract, simplified, generalized geometric, and detailed geometric descriptions, generated using $k$-shot prompting techniques \cite{brown2020language}.

In another recent work, Wang et al. \cite{wang2025text} used a VLM: llava-onevision-qwen2-7b \cite{li2024llava}, to generate initial draft captions for rendered CAD images, which were then refined through human annotations to create a dataset, pairing these captions with CAD parametric sequences. In \cite{wang2024cad}, Wang et al.~employed GPT-4o \cite{hurst2024gpt} to classify and filter CAD models, and then used InstructGPT \cite{wang2023self} to generate natural language descriptions. The resulting dataset contained CAD modeling sequences aligned with textual descriptions and rendered images from fixed viewpoints. Xie et al. \cite{xie2025text} built a dataset by converting minimal JSON from Text2CAD \cite{khan2024text2cad} into CadQuery code using Gemini 2.0 Flash \cite{pichai2024introducing}, with guidance and in-context examples. Code failures triggered self-correction \cite{madaan2023self}. Outputs were rendered via Blender \cite{blender41} and compared to ground truth. Samples were accepted only when both model and human agreed on shape similarity, yielding 170000 annotations. Yuan et al. \cite{yuan2025cad} generated synthetic data by leveraging VLMs. They created editing instructions by summarizing differences between the original and edited CAD models from both visual and sequence modalities, using GPT-4o \cite{hurst2024gpt} for the visual modality and LLaMA-3-70B \cite{grattafiori2024llama} for the sequence modality. Other examples of using LLM for CAD data generation include \cite{du2024blenderllm}, which developed BlendNet, a custom training dataset designed to capture diverse communication styles across three axes: 16 object categories, 8 instruction tones, and 5 length categories. Starting with 135 manually crafted seed instructions, the authors used self-instruct data distillation to expand to 50000 samples. GPT-4o \cite{hurst2024gpt} was used to generate scripts from these instructions, which were executed to produce corresponding images. GPT-4o then served as a validator to assess the alignment between images and instructions, resulting in a final dataset of 2000 human-verified and 6000 model-validated samples. Similarly, Li et al. \cite{li2024cad} also addressed the challenge of unavailable textual descriptions in existing CAD datasets. They applied the multimodal model CoCa \cite{yucoca} to generate captions for parametric CAD models in the DeepCAD dataset \cite{wu2021deepcad}, creating a new, richly annotated dataset linking text to CAD models. 

Likewise, in \cite{lv2025cadinstruct}, GPT-4o was employed to select CAD model names, while Claude-3.5-Sonnet was used to further refine, filter, and correct them. GPT-4o was also applied for image recognition of rendered views. DeepSeek-VL2 Small \cite{wu2024deepseekvl2mixtureofexpertsvisionlanguagemodels} generated comprehensive model descriptions, and DeepSeek-V3 \cite{liu2024deepseekv3} produced succinct explanations of the modeling logic for each component. In the same vein, Rosnitschek et al.~\cite{rosnitschek5555076dialogue} prompted Grok 3 to generate diverse multi-turn dialogues for constructing a bespoke dataset. Similarly, Govindarajan et al.~\cite{govindarajan2025cadmium} leveraged the multimodal capabilities of GPT-4.1 to generate concise, natural-sounding, yet geometrically precise descriptions, and constructed a new dataset. Guan et al.~\cite{guan2025cad} also prompted the code-generation LLM DeepSeek-V3 \cite{liu2024deepseekv3} to generate multiple candidate CadQuery scripts, selected the candidate with the lowest Chamfer Distance as the ground truth, and constructed a corresponding dataset. Moreover, in \cite{li2025seek}, the authors constructed a new CAD dataset of 40k samples following the Sketch, Sketch-based Feature, and Refinements modeling paradigm, with each sample paired with a textual description generated by GPT-4o.

It is worth noting that current research in data generation primarily leverages the text generation capabilities of LLMs. To date, in the CAD field, there is no work that directly generates other forms of LLM-based data, such as images, point clouds, or other format of data, to construct new datasets in the CAD domain.
\subsection{CAD Code Generation}
Many LLMs, such as Codex \cite{chen2021evaluating}, are pre-trained on vast code corpora, enabling them to generate code with a high degree of accuracy. Thus, a prominent application of LLMs in the CAD field is the generation of CAD code. In this direction, Li et al. \cite{li2024llm4cad, li2025llm4cad} explored the use of multimodal LLMs for 3D CAD generation, leveraging various data formats, including textual descriptions, images, sketches, and ground truth 3D shapes. In this work, LLMs were employed to generate CAD programs from these inputs, which were subsequently parsed into 3D shapes. Two models were evaluated: GPT-4 \cite{achiam2023gpt} and GPT-4V \cite{openai2023}. To improve the quality of the generated CAD code, the authors developed a debugger that iteratively refined the generated code until successful execution was achieved. The results showed that GPT-4V outperformed GPT-4, especially when using text-only input. It also surpassed other input combinations (e.g., text $+$ sketch, text $+$ image, and text $+$ sketch $+$ image) in terms of parsing rate and intersection over union. Sun et al. \cite{sun2025large} recently conducted a fine-tuning experiment on LLMs, comparing them to the baseline GPT-4 \cite{achiam2023gpt} without fine-tuning. They used the dataset from \cite{li2024llm4cad} and implemented four different sampling strategies for fine-tuning data generation. The first strategy randomly selected 60 code templates, the second ensured diversity in code length, the third focused on the shortest lengths, and the fourth on the longest. Overall, the second strategy achieved the highest average parsing rate, while the third performed best in terms of intersection over union. All four fine-tuned LLMs outperformed the baseline GPT-4 in most cases. In \cite{du2024blenderllm}, Du et al.~developed BlenderLLM, a script generation model that underwent SFT and iterative self-improvement. They fine-tuned the Qwen2.5-Coder-7B-Instruct model \cite{hui2024qwen2} using the BlendNet dataset and created a filter to select high-quality data generated by the model. This approach allowed for iterative optimization through cycles of data generation and model training. On similar lines, Alrashedy et al. \cite{alrashedygenerating} introduced CADCodeVerify, a framework designed to iteratively verify and refine 3D objects generated from CAD code. CADCodeVerify employs a VLM to assess the generated object by answering a set of questions and correcting deviations. They also introduced a benchmark for CAD code generation, containing 200 prompts paired with expert-annotated scripting code. The framework's feedback loop involved two steps: generating question-answer pairs and producing feedback without human intervention. The study compared the performance of three LLMs: GPT-4 \cite{achiam2023gpt}, Gemini \cite{team2023gemini}, and Code LLaMA \cite{roziere2023code}.

In \cite{DBLP:journals/corr/abs-2406-00144}, Badagabettu et al.~presented Query2CAD, a framework that generated CAD designs by using LLMs to produce executable CAD macros. The process involved the user query being fed into a robust LLM, such as GPT-3.5 Turbo \cite{openai2023gpt35turbo} or GPT-4 Turbo \cite{openai2023gpt4turbo}, which generated a Python macro. In cases where errors occurred, the error messages and Python code were provided to the LLM, which then generated corrected, executable code. The framework also utilized BLIP2 \cite{li2023blip} to generate captions for isometric views of CAD models, which were then passed onto the LLM to improve code generation through self-refinement loops. Yuan et al. \cite{YUAN2024104048} fine-tuned pre-trained LLMs to create their own LLM, called OpenECAD, which integrated the logical, visual, and coding abilities of VLMs. They rendered Boundary Representation (B-rep) data to generate 3D model images from three different views: default, orthographic, and transparent. For the base language models, they selected relatively small models like OpenELM \cite{mehta2024openelm}, Gemma \cite{team2024gemma}, and Phi-2 \cite{abdin2024phi3technicalreporthighly}. To endow OpenECAD with multimodal conversational capabilities, they leveraged training methods and datasets from LLaVA \cite{liu2023visual} and used GPT-4 \cite{achiam2023gpt} to generate multimodal image-language instruction data. The model was trained using the TinyLLaVA \cite{zhou2024tinyllavaframeworksmallscalelarge} framework and fine-tuned with the LoRA \cite{hu2022lora} method to enable the VLM to generate CAD code effectively. In addition, several studies have explored the potential of LLMs across a variety of design and manufacturing tasks. For instance, Makatura et al. \cite{Makatura2024Large, makatura2023can} evaluated the performance of LLMs in diverse design domains, including 2D vector graphic design (using SVG and DXF formats), 3D parametric modeling (using Constructive Solid Geometry (CSG) and B-rep formats), and articulated robotics problems (using the Universal Robot Description Format (URDF) and general graph-based representations). Except for URDF, the team initially used GPT-4 \cite{achiam2023gpt} to generate code, which was then converted into the corresponding 3D models in the desired format. In \cite{nelson2023utilizing}, GPT-4 \cite{achiam2023gpt} was used to generate CAD code, employing a back-and-forth dialogue process. Errors in the generated code and resulting structure were fed back into GPT-4, allowing it to iteratively refine the code. The study demonstrated GPT-4’s remarkable ability to generate unique solutions for constructing and debugging CAD code.

Mallis et al. \cite{mallis2024cad} utilized GPT-4o \cite{hurst2024gpt} to generate CAD code actions expressed in Python. The generated code was executed in CAD software, and the output was concatenated with the input context before being fed back to GPT-4o for the next iterative step. Rukhovich et al. \cite{rukhovich2024cad} proposed CAD-Recode, a solution for CAD reverse engineering. They fine-tuned an LLM to map input point clouds into CAD sketch-extrude sequences represented as Python code. The LLM used was Qwen2-1.5B \cite{yang2024qwen2technicalreport}, a relatively small model. CAD-Recode was augmented with a lightweight, trainable point cloud projector. The framework also utilized GPT-4o \cite{hurst2024gpt} to refactor the generated code in real time via interactive sliders. Jones et al. \cite{jones2025solver} worked on generative CAD design with GPT-4o \cite{hurst2024gpt} (without fine-tuning) and a solver-aided, domain-specific language: AI design language (AIDL). The process involved prompting GPT-4o with detailed descriptions of AIDL, including its syntax and available geometry types. The LLM then generated a complete AIDL program prompted by manually designed example programs, which was executed to create the desired model. If errors occurred during execution, they were fed back to GPT-4o, which would correct them.
Naik \cite{naik2024artificial} demonstrated the use of LLMs for creating extraction queries in SQL for geometry tagging, streamlining the tool's functionality by allowing users to provide natural language input instead of complex code. Four LLMs were compared: Gemini Pro \cite{team2023gemini}, GPT-3.5 Turbo \cite{openai2023gpt35turbo}, GPT-4, Claude 3 Sonnet \cite{TheC3}. Kienle et al. \cite{kienle2024querycad} designed QueryCAD, a deep learning-based question-answering system for CAD models. QueryCAD employed a code-writing LLM with in-context prompting to generate executable code for multi-view CAD part segmentation. It was integrated into MetaWizard, a robot program synthesis system \cite{alt2024robogrind}, to automatically generate industrial robot programs based on natural language task specifications. The team primarily used GPT-4o \cite{hurst2024gpt} and compared its performance with other LLaMA models (e.g., LLaMA-3 13B, LLaMA-3.1 8B, and LLaMA-3.1 405B) \cite{grattafiori2024llama}.

In \cite{ocker2025idea}, Ocker et al.~developed a VLM-based multi-agent system for CAD model generation. This system utilized multiple LLM-based agents to interpret sketches, images, and textual descriptions, generate CAD models, and verify the designs. The study proposed strategies for overcoming the well-known spatial reasoning limitations of VLMs, using GPT-4o \cite{hurst2024gpt} as the core LLM. In another recent work, Deng et al. \cite{DENG2024221} applied GPT-4 \cite{achiam2023gpt} to address three sub-tasks: parametric computation, instruction sequence construction, and coding model scripting. The process involved providing GPT-4 with design requirements, performing iterative design computations, and generating the necessary code by parsing a JSON file to formalize natural language instructions into executable commands. Doris et al. \cite{doris2025cad} introduced CAD-Coder, a fine-tuned open-source VLM for the generation of CAD code based on LLaVA-1.5 \cite{liu2023visual}. Given image inputs, the model produced readily executable Python CadQuery code. They compared their model against both closed-source and open-source VLMs, including GPT-o1 \cite{jaech2024openai}, GPT-4.5 \cite{openai2025gpt45}, and Gemini 2.0 \cite{pichai2024introducing} as closed-source, and InternVL2\_5-78B-MPO \cite{wang2024enhancing}, Ovis2-34B \cite{lu2024ovis}, and Qwen2.5-VL-72B \cite{qwen2.5-VL} as open-source. Xie et al. \cite{xie2025text} proposed generating CadQuery code directly from text using LLMs. They fine-tuned six open-source models of varying sizes, including CodeGPT-small \cite{microsoft2020codegpt}, GPT-2 medium and GPT-2 large \cite{radford2019language}, Gemma3-1B \cite{team2025gemma}, Qwen2.5-3B \cite{qwen2025qwen25technicalreport}, and Mistral-7B \cite{jiang2023mistral7b}. All models were fine-tuned via full-parameter supervised learning, except Mistral-7B, which adopted parameter-efficient fine-tuning using LoRA \cite{hu2022lora}. Kolodiazhnyi et al. \cite{kolodiazhnyi2025cadrille} proposed Cadrille, a CAD reconstruction model that jointly processed point clouds, images, and text, built on top of Qwen2-VL-2B \cite{wang2024qwen2}. They used a two-stage pipeline: SFT on large-scale procedurally generated data, followed by RL with online feedback, and showed that online RL outperformed offline alternatives.

Furthermore, in \cite{lv2025cadinstruct}, the authors employed Qwen2.5-Coder-7B to generate CAD programs. Likewise, Rosnitschek et al.~\cite{rosnitschek5555076dialogue} employed GPT-2 and CodeGen-350M for CAD code generation. In the same vein, Madireddy et al.~\cite{madireddy2025large} used BIM models and regulatory documents as input data and leveraged the capabilities of LLMs to generate Python scripts tailored for automated rule checking. Eight LLMs were evaluated: ChatGPT 4.0, Claude Sonnet 3.5, LLaMA 3.1, Microsoft Copilot, Gemini, Perplexity.AI, Grok, and DeepSeek. Similarly, in \cite{paviotiterative}, authors extracted B-Rep geometry from a CAD model and then used a dedicated API to generate a more concise and readable program. In addition, Campos et al.~\cite{campostext} tackled the text-to-SQL task for a specific type of real-world relational database containing data extracted from engineering CAD files. They introduced a task-specific prompt strategy and evaluated models including GPT-4o, LLaMA 3.1 8B, LLaMA 3.1 Instruct 70B, and Gemma 2 27B. Moreover, Chen et al.~\cite{chen2025cadreview} proposed a multimodal LLM-based framework consisting of a feedback generator and a code editor, which performed geometric operations by editing code for 3D object reconstruction. The framework employed two LLMs: Qwen2-VL \cite{wang2024qwen2} and LLaVA-OV \cite{li2024llava}. In \cite{guan2025cad}, Guan et al.~reformulated the text-to-CAD task as generating CadQuery code from natural language descriptions. Their approach consisted of two stages: supervised fine-tuning for CAD code generation and reinforcement learning with CAD-specific reward functions, using Qwen2.5-7B-Instruct \cite{qwen2025qwen25technicalreport} as the base model. Panta et al.~\cite{panta2025meda} also generated a Python script by importing the required libraries, which specified the parameters of the CAD model using the GPT-4o LLM.

\begin{figure*}
    \centering
    \includegraphics[width=\linewidth]{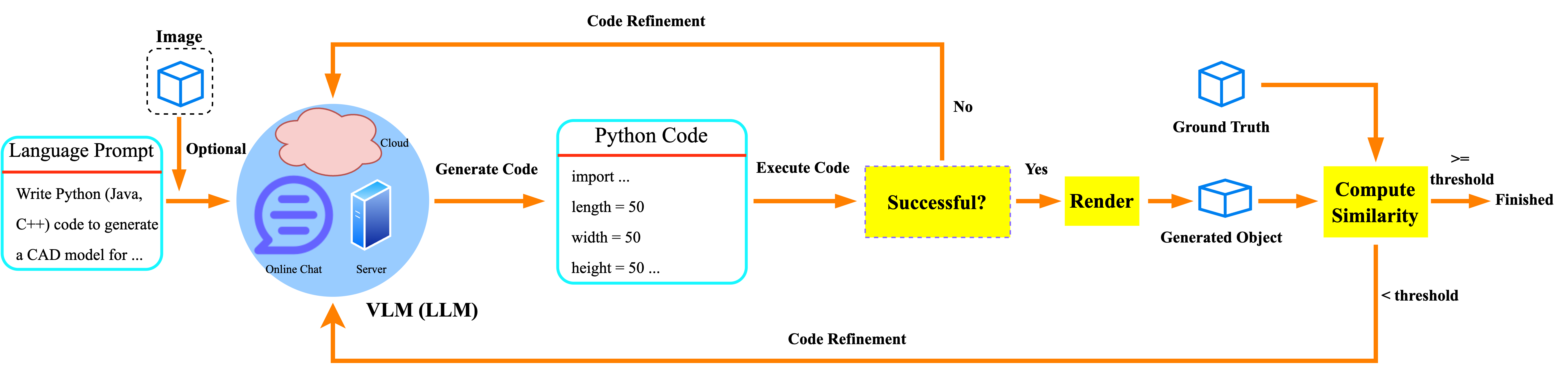}
    \caption{Typical CAD code generation pipeline. A prompt and, optionally, an image are first input to a VLM or LLM. The VLM/LLM then generates the corresponding code, which is executed. If the code is not executable, refinement is performed until it is. Subsequently, the generated objects' computing similarity to the ground truth is evaluated. If the similarity is below a threshold, the code refinement process is repeated until the desired results are achieved.
    \label{fig:codegeneration}}
\end{figure*}

CAD code generation remains one of the most crucial applications of LLMs in CAD. These generated code can take various forms, such as Python code, SQL code, CAD software-specific code, and even Java or C$++$ code. Users can generate the desired code formats based on their specific domain knowledge. Figure \ref{fig:codegeneration} illustrates the typical CAD code generation pipeline employed by most state-of-the-art approaches.
\subsection{Parametric CAD Generation}
Parametric CAD generation focuses on creating parametric data for CAD rather than code. In this application, LLMs generate parametric sequences, often formatted as JSON files. \cite{zhou2025status} provides a comprehensive review of multimodal LLMs for parametric CAD. In the direction of parametric CAD generation, Wu et al. \cite{wu2024cadvlm, wu2023cad} utilized pre-trained LLMs to efficiently manipulate engineering sketches, integrating sketch primitive sequences and images for parametric CAD generation. This included CAD autocompletion, CAD autoconstraint, and image-conditioned generation. They employed CodeT5+ \cite{wang-etal-2023-codet5} as the text encoder and decoder for sketch primitive sequences. To assess the general LLM's capability in CAD autocompletion, they fine-tuned ChatGPT on three subsets, demonstrating that ChatGPT outperformed other baselines. In  \cite{xu2025cadmllmunifyingmultimodalityconditionedcad}, Xu et al.~proposed CAD-MLLM, a system designed to generate parametric CAD models conditioned on multimodal inputs such as text, images, and point clouds. For images, they rendered multi-view images from eight fixed perspectives. For point clouds, they randomly sampled points at different ratios and recorded their corresponding normal information. They fine-tuned the open-source LLM Vicuna-7B \cite{chiang2023vicuna} and used LoRA \cite{hu2022lora} during training to minimize learnable parameters. You et al. \cite{you2024img2cad} focused on reverse engineering 3D CAD models from images. They first used the GPT-4V \cite{openai2023} foundation model to predict a global discrete base structure, extracting semantic information from images. The model identified semantic parts from the image before generating CAD sequences for each part. They then built a Transformer model to predict continuous attribute values based on the discrete structure with semantics.

Zhang et al. \cite{zhang2025flexcad} applied LLMs for controllable generation across various CAD construction hierarchies. They initially converted a CAD model into structured text. They then fine-tuned an LLM, LLaMA-3-8B \cite{grattafiori2024llama}, to develop a unified model for controllable CAD generation. During training, a hierarchy-aware field in the CAD text was masked, and LLMs were tasked with predicting the masked field. During inference, users could specify the part to modify by masking it and inputting it into the LLM for generating new CAD models. Wang et al. \cite{wang2024cad} proposed a CAD synthesis method using a multimodal LLM with enhanced spatial reasoning capabilities. They input either a single image or text and mapped the 3D space into 1D using a tokenization method to improve spatial reasoning. They employed LLaVA-1.5 7B \cite{Liu_2024_CVPR} as their base model, with pre-trained Vicuna \cite{chiang2023vicuna} as their foundation, which was built on LLaMA-2 \cite{touvron2023llama2}. The 3D CAD model was represented in JSON format, capturing key modeling commands and parameters in the order of CAD construction, based on the DeepCAD dataset \cite{wu2021deepcad}. Utilizing the LLaMA-3 8B Instruct \cite{grattafiori2024llama} LLM as the backbone,  Wang et al. \cite{wang2025text} introduced a CAD-Fusion framework. It alternated between two training stages: the sequential learning stage, which trained LLMs using ground truth parametric sequences, and the visual feedback stage, which rewarded parametric sequences that rendered visually preferred objects and penalized those that did not. Makatura et al. \cite{Makatura2024Large, makatura2023can} created a design space defined by parametric design and the bounds of these parameters, encompassing a range of potential designs. When prompted with lower and upper bounds for parameters, LLMs suggested values based on typical proportions for the designed object. While the absolute scale was arbitrary, the proposed bounds were semantically reasonable and proportionate. In addition to creating design spaces, they also used GPT-4 \cite{achiam2023gpt} to generate XML-format data rather than code, focusing on pre-existing designs in formats like URDF. Li et al. \cite{li2025cad} proposed CAD-Llama, a comprehensive framework that adapted open-source LLMs for generating CAD sequences. They introduced a code-like format to translate parametric 3D CAD command sequences into structured parametric CAD code. For training, they employed an adaptive pretraining approach combined with instruction tuning for various downstream tasks, which enabled the LLM to acquire CAD modeling capabilities and adapt to diverse applications, selecting LLaMA3-8B \cite{grattafiori2024llama} as the foundational LLM. Yuan et al. \cite{yuan2025cad} introduced a text-based CAD editing framework, where the input consisted of an editing instruction and the sequence representation of the original CAD model, and the output was the sequence of the edited model. This approach enabled precise modifications through natural language to better reflect real-world user intent. They fine-tuned LLaMA-3-8B-Instruct \cite{grattafiori2024llama} using LoRA.

Likewise, in \cite{ghosh2025fostering}, Gosh et al.~represented customer requirements as text- or sketch-based descriptions of the product to be manufactured and delivered by the manufacturer. Furthermore, Yin et al.~\cite{yin2025shapegpt} introduced ShapeGPT, an LLM trained to produce shape token sequences from text, images, or multimodal inputs. The model also supports downstream tasks including shape editing, shape reasoning, and shape completion. Similarly, the authors of \cite{daareyni2025generative} proposed a framework for LLM-assisted CAD using advanced AI systems, specifically the GPT-4o model, enabling users to generate and manipulate 3D models via text commands, visual inputs (e.g., blueprints, images), voice, and other types of general or professional information. Moreover, Rietschel et al.~\cite{rietschel2025intelligent} outlined three principles for developing LLM-driven tools in design research and demonstrated these principles through a system capable of generating parametric design scripts via reasoning-enhanced models. In parallel, in \cite{zhanggeocad}, GeoCAD took three inputs: an original CAD model represented in a hierarchical textual format, a selected local part, and user-provided geometric instructions, and generated a new CAD model that modified only the designated local part while closely adhering to the given instructions. In the same vein, Govindarajan et al.~\cite{govindarajan2025cadmium} fine-tuned Qwen 2.5-Coder-14B on their dataset to generate JSON-formatted CAD sequences from natural language prompts, effectively leveraging the pre-trained model’s capabilities without introducing any specialized embedding layers. In \cite{li2025seek}, Li et al.~also leveraged the locally deployed open-source inference LLM DeepSeek-R1 for parametric CAD model generation. The initially generated parametric model was rendered into a sequence of step-wise perspective images, which were then processed by a VLM together with the corresponding CoT traces produced by DeepSeek-R1 to assess and subsequently refine the CAD model. Analogously, in \cite{mccarthy2025mrcad}, the authors proposed a novel task for studying and benchmarking multimodal refinement in a CAD design setting. One player, the designer, had a target design which they demonstrated via a combination of text and drawing instructions. The other player, the maker, used a programmatic CAD language to carry out the designer's instructions and attempted to recreate the target design. They had several LLMs, including GPT-4o, GPT-4o-mini, Claude-3.7-Sonnet, and Qwen2.5-VL-7B-Instruct \cite{bai2025qwen2}, assume the role of the human maker. In a similar manner, Panta et al.~\cite{panta2025meda} interpreted and analyzed the user-provided specifications and generated a modeling plan outlining the relevant methods and functions of the open-source Python library.

\begin{figure*}
    \centering
    \includegraphics[width=\linewidth]{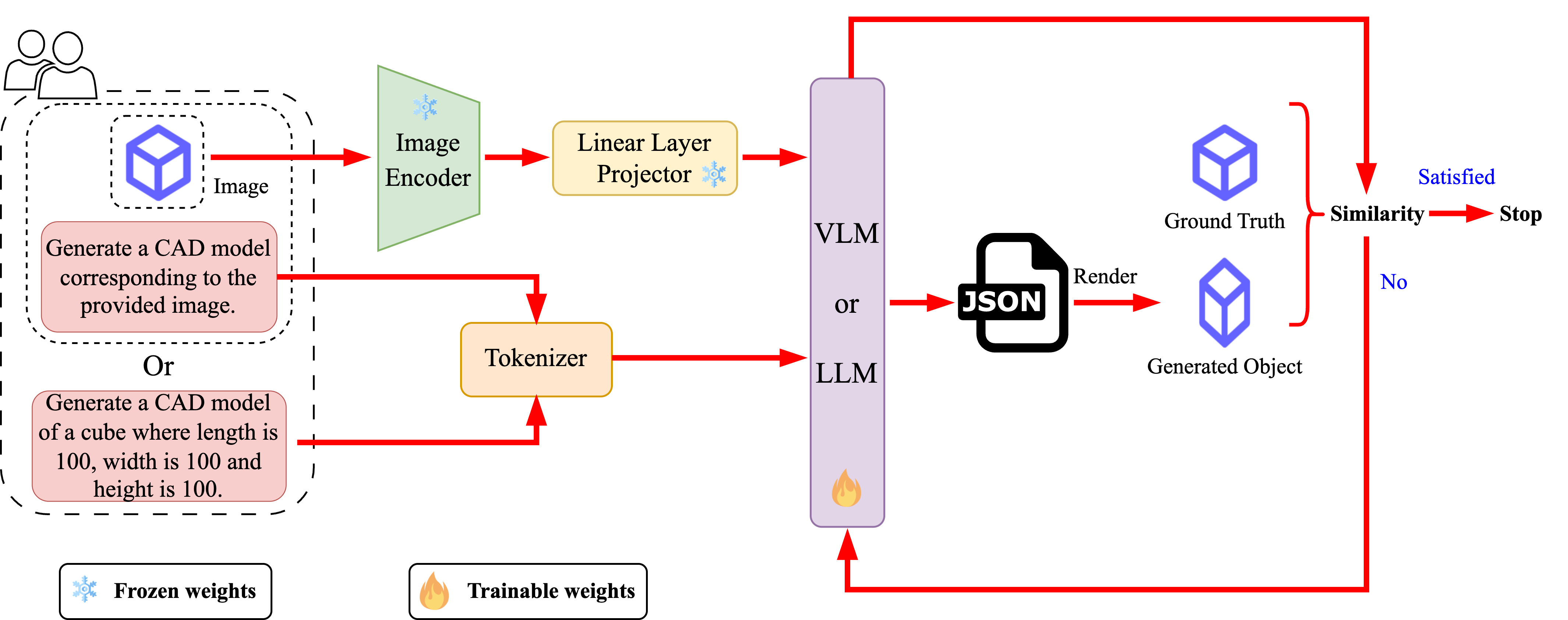}
    \caption{A standard parametric CAD generation pipeline. A prompt, along with an optional image (converted into features by a frozen image encoder), is fed into a trainable VLM or LLM to generate parametric data. This data is then parsed to produce 3D CAD models. LLMs can also be employed to compute the similarity between the ground truth and the generated object.
    \label{fig:parametriccadgeneration}}
\end{figure*}

Parametric CAD Generation is another key application of LLMs in the CAD field. Unlike CAD code generation, which relies on LLMs to generate executable code, parametric CAD generation uses LLMs to produce parametric sequences instead of code. While generating accurate and executable code can be challenging for state-of-the-art LLMs, parametric CAD generation is often preferred by researchers. Although LLMs still face difficulties in producing perfectly accurate parametric sequences, if the goal is to generate keys and values rather than fully executable sequences, many CAD challenges can be alleviated. Figure \ref{fig:parametriccadgeneration} illustrates a standard parametric CAD generation pipeline used by most state-of-the-art methods in this direction. 
\subsection{Image Generation}
Whereas generative technologies are becoming increasingly popular in visual modeling~\cite{ulhaq2022efficient}, the use of LLMs in image generation tasks related to CAD is still a widely open direction. Recently,  Tang et al. \cite{tang2025chatcad} employed LangGraph\footnote{\url{https://www.langchain.com/built-with-langgraph}}, an LLM, to perform digital CAD drawing restoration and utilized Retrieval-Augmented Generation (RAG) \cite{lewis2020retrieval} technology to incorporate engineering domain knowledge into the model. This is the only work we currently find that generates CAD drawings from an LLM in an end-to-end manner.
\subsection{Model Evaluation}
LLMs can also be used to compare two sources of data. For instance, ChatGPT can be used to estimate similarity between two textual descriptions: one generated by a model and the other being the ground truth. This ability of LLMs is also making its way into the CAD domain. For instance, Du et al. \cite{du2024blenderllm} developed an evaluation framework, CADBench, which utilized GPT-4o \cite{hurst2024gpt} for two complementary evaluation tasks: image-based evaluation, which assessed visual fidelity, and script-based evaluation, which compared objective attributes such as size, color, and material. In \cite{DBLP:journals/corr/abs-2406-00144}, the visual question-answering score between the user query and the generated isometric image was computed using Clip-FlanT5-XL \cite{lin2024evaluating}. Similarly, in \cite{tang2025chatcad}, Tang et al.~fed repaired drawings alongside ground truth drawings into GPT-4 \cite{achiam2023gpt} for automatic scoring along three dimensions: legibility, completeness, and tolerance. Wang et al. \cite{wang2025text} leveraged the strong visual understanding capabilities of VLMs to score visual objects, aiding in the construction of preference data, which would otherwise be costly and labor-intensive. The rendered CAD images, along with an instruction detailing the evaluation criteria divided into three categories: shape quality, shape quantity, and distribution, were passed into the VLM, LLaVA-OV-Qwen2-7B \cite{li2024llava}, to compute the scores.
\subsection{Text Generation}
Text generation is one of the most common applications of LLMs. Unsurprisingly, many CAD-related works have also employed LLMs for generating text content. For example, Liu et al. \cite{liu20233dall} combined DALL-E \cite{ramesh2022hierarchical}, GPT-3 \cite{brown2020language}, and CLIP \cite{radford2021learning} within CAD software, enabling users to construct text and image prompts based on their modeling needs. Designers could use text-to-image AI to generate reference images, avoid design bias, prevent design mindset, and stimulate new design considerations. Users would input their design intentions, and GPT-3 would suggest prompts for them to select from. After choosing, the suggestions were rephrased, and DALL-E would generate the final results. Kodnongbua et al. \cite{10.1145/3610548.3618219} employed ChatGPT to generate text prompts describing variations of a given model. Similarly, Yuan et al. \cite{Yuan_2024_CVPR} introduced a task for semantic commenting of CAD programs. In this task, LLMs segmented programs into code blocks, each corresponding to semantically meaningful parts, and assigned semantic labels to each block. They executed the program to generate a 3D shape and rendered images from ten representative viewpoints. Each image was then translated into a photorealistic image using ControlNet \cite{zhang2023adding}. Using ChatGPT, the authors segmented the images into semantically meaningful parts and transformed every bounding box to a pixel-wise segment utilizing segment anything model \cite{Kirillov_2023_ICCV}, linking them to corresponding code blocks. They also explored how ChatGPT could comment on programs based on a given example program, finding that while it performed well at commenting similar programs, it struggled to generalize to new shapes and primitives.

In \cite{you2024img2cad}, You et al. employed a VLM to generate semantic part comments for their predicted global base structure. They used GPT-4V \cite{openai2023} to interpret an input image and decompose it into semantic parts. Mallis et al. \cite{mallis2024cad} fed multimodal user requests into a VLM, which then outputted a response in the form of a natural language plan. They compared three VLMs: GPT-4o \cite{hurst2024gpt}, GPT-4 mini and GPT-4 Turbo \cite{openai2023gpt4turbo}. Along similar lines, Rukhovich et al. \cite{rukhovich2024cad} utilized GPT-4o \cite{hurst2024gpt} for CAD-specific question answering, allowing users to query the system for information related to CAD models. In another recent effort, Yu et al. \cite{yu2024intelligent} employed a locally built LLM to analyze users' requirements and generate initial design ideas. The LLM was used to establish structural requirements and guide how to model the scheme. They also employed finite element analysis to improve data and fine-tuned the LLM to generate easily understandable design improvement suggestions. The system used the generalv3.5 model \cite{iflytekgeneralv3.5} provided by the iFlytek AI platform. An investigation of VLMs for automatically recognizing manufacturing features in CAD design was conducted in \cite{khan2024leveraging}. The process began by converting a CAD file into three different image views, which, combined with prompts outlining the analysis criteria, served as inputs for five VLMs to identify and evaluate the features. The VLMs included both closed-source models such as GPT-4o \cite{hurst2024gpt}, Claude-3.5-Sonnet \cite{anthropic2024claude35sonnet}, and Claude-3-Opus \cite{TheC3}, as well as open-source models like MiniCPM-Llama3-V2.5 \cite{hu2024minicpm} and Llava-v1.6-mistral-7b \cite{liu2024llavanext}. A review along conceptually similar lines was also conducted by Sun et al. \cite{sun2024ai}, who analyzed key developments in rule-based reasoning (RBR), case-based reasoning (CBR), and large AI models for advancing reusable design in CAD software, summarizing both their advantages and disadvantages. They proposed a hybrid framework combining RBR, CBR and LLMs along with knowledge graphs to enhance reusable CAD design. They mentioned several LLMs, including GPT-4 \cite{achiam2023gpt}.

Picard et al. \cite{picard2023concept} explored the capabilities of GPT-4V \cite{openai2023} and LLaVA 1.6 34B \cite{liu2024llavanext} in performing various engineering design tasks, which involved both visual and textual information. These tasks were categorized into four main areas: conceptual design, system-level and detailed design, manufacturing and inspection, and engineering education. Most of these tasks involved generating textual descriptions from images and text prompts. They also assessed GPT-4V's spatial reasoning abilities, concluding that while the model demonstrated some spatial reasoning capability, it was still limited when compared to human performance. In \cite{CAD2Program}, Wang et al. treated 2D drawings as raster images and employed an image encoder to extract features. They utilized general-purpose language scripts to represent 3D parametric models and leveraged an LLM to autoregressively predict parametric sequences in text form. They developed CAD2Program by fine-tuning an open-source VLM, Mini-InternVL-1.5-2B \cite{gao2024mini}, which used InternViT-300M \cite{chen2024expanding} as the vision encoder and InternLM2-1.8B \cite{cai2024internlm2} as the language model. Relevant applications of LLMs in manufacturing were reviewed by Li et al.~in \cite{li2024large}, evaluating LLM use in various tasks through case studies and examples. While this survey also covered areas outside CAD, it discussed CAD-related tasks including data generation, text-grounded 3D content generation, initial design drafts, and idea generation in aerospace design. They mentioned GPT-4V \cite{openai2023} and OpenAI’s o1 model \cite{jaech2024openai} as important LLMs for CAD in their paper. The study in \cite{10.1115/1.4067318} examined the impact of ChatGPT-3.5 \cite{openai_gpt35} on engineering design education, focusing on concept generation and detailed modeling. It found that while the LLM significantly broadened concept diversity, its textual nature and occasional unreliable outputs limited its effectiveness in detailed CAD modeling, underscoring the irreplaceable value of traditional learning resources and hands-on practice.

Similarly, Garcia et al.~\cite{garcia2025llms} leveraged Qwen-2, a pre-trained LLM, for a feature classification task in which ASCII-formatted STL files were treated as language-based data, but the approach failed due to repeated interruptions and out-of-memory errors. Likewise, Yin et al.~\cite{yin2025shapegpt} also trained ShapeGPT to perform shape captioning, enabling natural-language descriptions of 3D shapes. In parallel, in \cite{lee5396057automated}, the authors leveraged LLMs to perform material-layer function assignment and building-object type classification. In the same vein, Doris et al.~\cite{doris2025designqa} projected 3D CAD models into 2D through three forms of imagery: multiview CAD renderings, close-up detail views, and engineering drawings. They further introduced DesignQA, a benchmark organized into three components: rule extraction, comprehension, and compliance. They evaluated five LLMs: GPT-4o, GPT-4, Gemini-1.0, Claude-Opus, and LLaVa-1.5. Furthermore, Li et al.~\cite{li2025automated} performed knowledge analysis based on the semantic understanding capabilities of LLMs and generated a set of triples comprising six components: the function set, structure set, constraint set, function–structure relationships, structure–structure relationships, and structure–constraint relationships. Moreover, in \cite{hu2025question}, entities, attributes, and relations are extracted from each CAD model and the corresponding graph structure is computed to obtain a vector representation. Then,  Vicuna is employed to parse natural language queries and to perform question answering. In addition, Zhang et al.~\cite{zhanggeocad} captioned $\sim$221k local parts, using vertex-based captioning for simple parts and VLM-based captioning for more complex ones. Authors of \cite{valente2025cad2dmd} also evaluated and fine-tuned three VLMs on their visual question answering (VQA) performance, Pixtral-12B \cite{agrawal2024pixtral}, LLaVA, and InternVL, all of which operated on image inputs. In a similar manner, Niu et al.~\cite{niu2025creft} used a ViT visual encoder and Qwen2.5 for CAD-based VQA tasks. Similarly, \cite{manvideocad} evaluated a range of multimodal LLMs on spatial reasoning and video understanding tasks derived from CAD designs, including gpt-4.1-2025-04-14, claude-3-7-sonnet-20250219, qwen2.5-vl-72b-instruct \cite{bai2025qwen2}, o3-2025-04-16, and gemini-2.5-pro-preview-05-06. In a like manner, Panta et al.~\cite{panta2025meda} analyzed the most recently generated CAD model image after execution, produced a description using GPT-4o, and compared it with the design prompt to evaluate correct model generation.

Text generation is primarily associated with question answering. While this application may seem simple, it is essential to recognize that it is one of the key strengths of LLMs. By effectively leveraging the text and content generation capabilities of LLMs, we can achieve results that go beyond initial expectations, unlocking significant potential for innovative solutions.
\section{CAD Evaluation}\label{sec:eval}
To evaluate the capability of their  methods, existing  studies have introduced a range of assessment metrics. For instance, You et al.~\cite{you2024img2cad} evaluated models using Chamfer Distance, part-segmentation accuracy, and part-segmentation mIoU. Similarly, evaluation in \cite{rukhovich2024cad} also used Chamfer Distance, Intersection over Union, and Invalidity Ratio. Likewise, Mallis et al.~\cite{mallis2024cad} employed 2D and 3D Accuracy, and for autoconstraining evaluated performance using Primitive F1 and Constraint F1. They also used Chamfer Distance from \cite{khan2024text2cad} and additional metrics from \cite{xu2022skexgen}, including Coverage, Minimum Matching Distance, and Jensen–Shannon Divergence, along with Invalidity Ratio. They also introduced a VLM-based metric termed the LVM score. Moreover, Xie et al.~\cite{xie2025text} used Chamfer Distance~\cite{fan2017point}, Invalid Rate, and Gemini 2.0 Flash~\cite{wu2024gpt, zhang2023gpt} as evaluation metrics. Following \cite{rukhovich2024cad}, \cite{kolodiazhnyi2025cadrille} also evaluated Chamfer Distance, Intersection over Union, and Invalidity Ratio. In addition, Wang et al.~\cite{wang2024cad} used Chamfer Distance~\cite{fan2017point}, Invalidity Ratio, Command Accuracy, and Parameter Accuracy. Furthermore, \cite{lv2025cadinstruct} evaluated code executability, geometric similarity (Chamfer Distance), and image similarity using DINOv2~\cite{oquab2023dinov2}. Similarly, in \cite{li2025cad}, unconditional CAD generation was evaluated using Coverage, Minimum Matching Distance, Jensen–Shannon Divergence, success ratio $S_R$, and Novel score. For text-to-CAD, the authors used reconstruction accuracy $ACC_T$~\cite{li2024cad}, computed with command accuracy $ACC_{cmd}$, parameter accuracy $ACC_{param}$~\cite{wu2021deepcad}, and success ratio $S_R$. Median Chamfer Distance, Minimum Matching Distance, and Jensen–Shannon Divergence were also employed. CAD captioning was assessed with BLEU~\cite{papineni2002bleu} and ROUGE~\cite{lin2004rouge}. CAD deletion used Extract Match, while CAD addition used $ACC_{cmd}$ and $ACC_{param}$. Likewise, \cite{li2025seek} used Chamfer Distance, Hausdorff Distance, and IoGT, G-Score and proposed a novel metric. Guan et al.~\cite{guan2025cad} used mean Chamfer Distance, median Chamfer Distance, and Invalidity Ratio as their metrics. 

In \cite{xu2025cadmllmunifyingmultimodalityconditionedcad}, authors evaluated reconstruction with Chamfer Distance, F-score, and Normal Consistency, topology with Segment Error, Dangling Edge Length, and Self-Intersection Ratio, and assessed enclosure with Flux Enclosure Error. Moreover, \cite{li2024cad} adopted a self-designed accuracy along with Median Chamfer Distance, Minimum Matching Distance, Jensen–Shannon Divergence, and CT-Score. Furthermore, \cite{wang2025text} adopted the F1 score following \cite{khan2024text2cad}, reporting F1-Sketch and F1-Extrusion. In addition, Liu et al.~\cite{liu20233dall} quantified how frequently participants used LLM-provided prompt suggestions. Similarly, \cite{wu2024cadvlm} introduces sketch accuracy, entity accuracy, and CAD F1 as metrics. In the same vein, Alrashedy et al.~\cite{alrashedygenerating} used Point Cloud Distance, Hausdorff Distance, and Intersection over the Ground Truth (IoGT), while  \cite{khan2024leveraging} adopted Feature Name Accuracy, Feature Quantity Accuracy, Hallucination Rate, and Mean Absolute Error. \cite{zhang2025flexcad} adopted metrics from \cite{xu2022skexgen, xu2023hierarchical}, including Coverage, Minimum Matching Distance, and Jensen–Shannon Divergence. Inspired by \cite{wu2024gpt, he2023t}, \cite{khan2024text2cad} also used F1 scores of primitives and extrusions, Chamfer Distance, and Invalidity Ratio, and additionally performed GPT-4V and user-study evaluations, both measured by accuracy. Likewise, Yin et al.~\cite{yin2025shapegpt} used 3D Intersection over Union, Chamfer Distance, F-score@1\%~\cite{10.1007/11941439_114}, and ULIP similarity~\cite{xue2023ulip}. In \cite{govindarajan2025cadmium}, the authors adopt Invalidity Ratio, F1, Chamfer Distance, Segment Error, Dangling Edge Length, Self-Intersection Ratio, Flux Enclosure Error, and Sphericity Discrepancy, Discrete Mean Curvature Difference, and Extract Euler Characteristic Match. Furthermore, in~\cite{yuan2025cad}, for text-based CAD editing, the authors evaluated Valid Ratio, Jensen--Shannon Divergence following \cite{wu2021deepcad, xu2022skexgen, xu2023hierarchical}, and Chamfer Distance as \cite{khan2024text2cad}. For evaluating instruction alignment, they adopt Directional CLIP Score from \cite{gal2022stylegan, brooks2023instructpix2pix}. On the other hand, Yuan et al.~\cite{YUAN2024104048} designed a scoring algorithm that checks whether the generated code can be executed and then uses GPT-4o to compare the rendered model view with the target input. In \cite{Yuan_2024_CVPR}, the authors proposed block accuracy and semantic Intersection over Union (IoU) to evaluate models on commenting CAD programs. Similarly, Badagabttu et al.~\cite{DBLP:journals/corr/abs-2406-00144} reported success rate, and \cite{kienle2024querycad} reported accuracy. Evaluation in \cite{doris2025cad} employed Valid Syntax Rate and $\mathrm{IOU}_{\mathrm{best}}$. Likewise, Rosnitschek et al.~\cite{rosnitschek5555076dialogue} measured code-generation error, validation loss, and training time. In \cite{li2024llm4cad}, Li et al.~employed parsing rate and intersection-over-union, while \cite{li2025llm4cad, sun2025large} also used the same metrics. However, Garcia et al.~\cite{garcia2025llms} reported F1 score, training time, GPU prediction time, and CPU prediction time. In addition, \cite{doris2025designqa} employed F1 score, accuracy, BLEU, ROUGE, and similarity. Likewise, \cite{madireddy2025large} evaluated processing time, number of correction attempts, a binary status indicator, and overall success rate. In \cite{jones2025solver}, authors adopted the average CLIP score as metric.  Similarly, Chen et al.~\cite{chen2025cadreview} employed ROUGE$_L$ \cite{lin2004rouge}, BERTScore \cite{zhangbertscore}, and accuracy. Following \cite{haque2022deepcad}, they further used Chamfer Distance, Minimum Matching Distance, Jensen–Shannon Divergence, and Invalidity Ratio. \cite{10.1145/3610548.3618219} measured precision and recall. In contrast, Picard et al.~\cite{picard2023concept} employed several evaluation metrics: self-consistency, transitive violations, accuracy, volume fraction error, floating material error, topology optimization metrics, force magnitude, volume fraction VF, filter radius R, and load application angle. Differently, Li et al.~\cite{li2025automated} took accuracy, time, and valid triplets as their evaluation metrics. However, \cite{hu2025question} employed query accuracy, question-answering accuracy, and F1-score. On the other hand, \cite{campostext} assessed the models in terms of parameters, duration, number of correct and incorrect answers, and overall accuracy. Conversely, Zhang et al.~\cite{zhanggeocad} evaluated their method using Coverage, Minimum Matching Distance, Jensen–Shannon Divergence, and prediction validity as in \cite{zhang2025flexcad}, supplemented by vertex-based and VLM-based scores, as well as a realism metric reflecting human judgments. Likewise, Panta et al.~\cite{panta2025meda} employed point cloud distance, Hausdorff distance, and IoGT as evaluation criteria. Nevertheless, Sadik et al.~\cite{sadik2025human} introduced a composite complexity score to quantify structural complexity, and a final similarity score to assess structural fidelity. In a different vein, Agarwal et al.~\cite{agrawal2024pixtral} used Average Normalized Levenshtein Similarity \cite{peer2024anls} as the primary evaluation metric, alongside training data usage, word-level accuracy, unit accuracy, and numeric accuracy. In comparison, Wang et al.~\cite{CAD2Program} evaluated model retrieval accuracy, 3D reconstruction quality, and parameter estimation accuracy. For 3D reconstruction, they report precision, recall, and F1 score. For model retrieval, they assess the match percentage, and for parameter estimation, accuracy over correctly retrieved models is reported. Alternatively, \cite{mccarthy2025mrcad} evaluated the proportional improvement. From another standpoint, Du et al.~\cite{du2024blenderllm} computed the evaluation metric by averaging the outputs across all criteria. However, Tang et al.~\cite{tang2025chatcad} adopted legibility, completeness, and tolerance, as well as PSNR and SSIM from industrial drawing evaluation. Nonetheless, \cite{manvideocad} employed Command and Parameter Accuracy, Offline Closed-Loop Execution Performance, and Geometric Fidelity.

\pgfplotsset{compat=1.18}

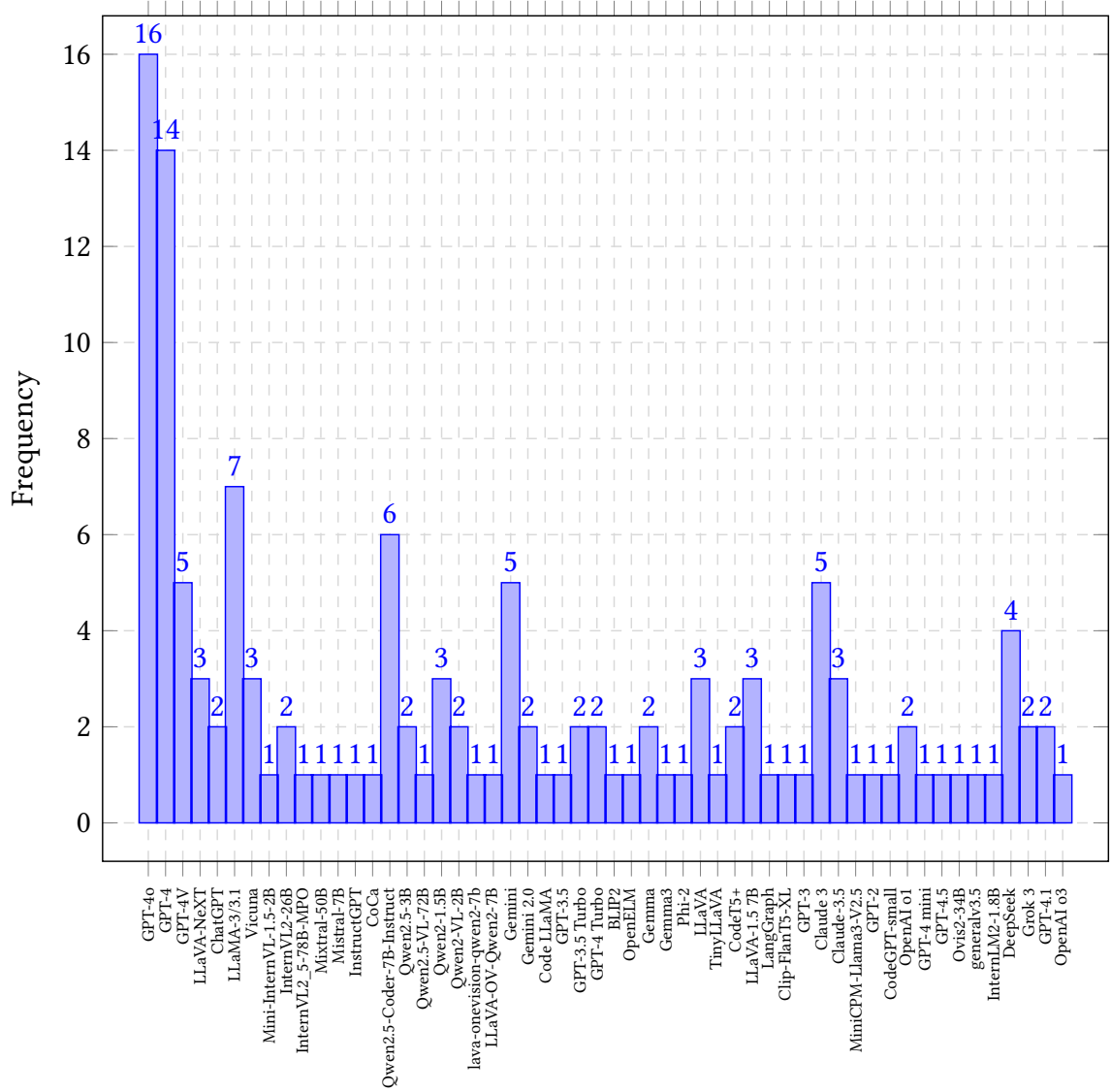
\begin{figure}[t]
\centering
\begin{adjustbox}{width=\textwidth}
\begin{tikzpicture}
\begin{axis}[
    ybar,
    bar width=5.2pt,
    enlargelimits=0.05,
    ylabel={Frequency},
    symbolic x coords={
        GPT-4o, GPT-4, GPT-4V, LLaVA-NeXT, ChatGPT, LLaMA-3/3.1, Vicuna,
        Mini-InternVL-1.5-2B, InternVL2-26B, InternVL2\_5-78B-MPO, Mixtral-50B, Mistral-7B, InstructGPT, CoCa,
        Qwen2.5-Coder-7B-Instruct, Qwen2.5-3B, Qwen2.5-VL-72B, Qwen2-1.5B, Qwen2-VL-2B, lava-onevision-qwen2-7b, 
        LLaVA-OV-Qwen2-7B, Gemini, Gemini 2.0, Code LLaMA, GPT-3.5, GPT-3.5 Turbo, GPT-4 Turbo, 
        BLIP2, OpenELM, Gemma, Gemma3, Phi-2, LLaVA, TinyLLaVA, 
        CodeT5+, LLaVA-1.5 7B, LangGraph, Clip-FlanT5-XL, 
        GPT-3, Claude 3, Claude-3.5, MiniCPM-Llama3-V2.5,
        GPT-2, CodeGPT-small, OpenAI o1, GPT-4 mini, GPT-4.5, Ovis2-34B, generalv3.5, InternLM2-1.8B, DeepSeek, Grok 3, GPT-4.1, OpenAI o3
    },
    xtick=data,
    xticklabel style={rotate=90, anchor=east, font=\tiny},
    nodes near coords,
    ymin=0,
    height=10cm,
    grid=both,
    major grid style={dashed,gray!30},
    minor grid style={dotted,gray!20}
]
\addplot coordinates {
    (GPT-4o,16) (GPT-4,14) (GPT-4V,5) (LLaVA-NeXT,3) (ChatGPT,2) (LLaMA-3/3.1,7) (Vicuna,3)
    (Mini-InternVL-1.5-2B,1) (InternVL2-26B,2) (InternVL2\_5-78B-MPO,1) (Mixtral-50B,1) (Mistral-7B,1) (InstructGPT,1) (CoCa,1) 
    (Qwen2.5-Coder-7B-Instruct,6) (Qwen2.5-3B,2) (Qwen2.5-VL-72B,1) (Qwen2-1.5B,3) (Qwen2-VL-2B,2) (lava-onevision-qwen2-7b,1)
    (LLaVA-OV-Qwen2-7B,1) (Gemini,5) (Gemini 2.0,2) (Code LLaMA,1) (GPT-3.5,1) (GPT-3.5 Turbo,2) (GPT-4 Turbo,2)
    (BLIP2,1) (OpenELM,1) (Gemma,2) (Gemma3,1) (Phi-2,1) (LLaVA,3) (TinyLLaVA,1)
    (CodeT5+,2) (LLaVA-1.5 7B,3) (LangGraph,1) (Clip-FlanT5-XL,1)
    (GPT-3,1) (Claude 3,5) (Claude-3.5,3) (MiniCPM-Llama3-V2.5,1)
    (GPT-2,1) (CodeGPT-small,1) (OpenAI o1,2) (GPT-4 mini,1) (GPT-4.5,1) (Ovis2-34B,1) (generalv3.5,1) (InternLM2-1.8B,1) (DeepSeek,4) (Grok 3,2) (GPT-4.1,2) (OpenAI o3,1)
};
\end{axis}
\end{tikzpicture}
\end{adjustbox}
\caption{Bar chart of LLM usage frequency in CAD-related research.}
\label{fig:barchart}
\end{figure}
\section{Analysis and Discussion}\label{analysis}
As  clear from Section~\ref{sec:eval}, commonly adopted metrics  for CAD evaluation include accuracy, Chamfer Distance, success rate, Valid/Invalid ratio, Minimum Matching Distance, Jensen--Shannon Divergence, Coverage, F1 score, and Intersection over Union. These metrics cover a range of objectives popular for CAD tasks. Furthermore, from the aforementioned state-of-the-art works in Section \ref{application}, it can be concluded that nearly all approaches utilize LLMs to generate intermediate representations rather than directly outputting 3D CAD models or 2D CAD drawings. This is likely because directly generating accurate 3D or 2D CAD outputs remains a significant challenge for current models. Common intermediate formats include executable code, such as Python scripts, and parametric data, often structured as JSON files, which can then be parsed or executed to construct the final CAD models. We also make another key observation that most state-of-the-art methods rely on multimodal inputs, making the use of multimodal LLMs increasingly prevalent. This highlights the potential of different input types for LLM-based automation. Among these input types, text and images are used most frequently. However, other data formats such as point clouds and sketch sequences are also explored, reflecting the growing diversity of input modalities being considered in the field. We also record the frequency with which each LLM is used in the state-of-the-art works that we review. Figure \ref{fig:barchart} summarizes the number of times different LLMs have been employed in the reviewed contributions. As apparent, GPT-4o and GPT-4 are the most frequently used models (16 and 14 instances respectively), all developed by OpenAI. This trend suggests that, despite being closed-source and potentially requiring payment, LLMs from the GPT family are still widely favored by the research community in CAD domain. A likely reason is their superior performance compared to the other types of LLMs, including many publicly available models. We also analyze the distribution of reviewed works across different application types. Table \ref{tab:number} summarizes the number of studies attributed to each category. As shown, the majority of contributions focuses on CAD code generation (27 works), followed by text generation (24 works) and parametric CAD generation (20 works). This suggests that CAD code generation is currently the most prominent and attractive application of LLMs in the CAD domain, and it is expected to continue gaining traction among researchers. Text generation, being a core capability of LLMs, also remains a widely explored area within the research community. We also analyze the datasets used in each state-of-the-art study. Table \ref{tab:dataset} presents the datasets employed across the reviewed works. Some datasets are sourced from industry data, while others are synthetically generated by the authors. Many of the industrial datasets are derived from publicly available datasets, such as DeepCAD, Onshape\footnote{\url{http://onshape.com}}, ABC \cite{koch2019abc}, and others. Finally, we map each state-of-the-art work to its industrial context. This allows us to identify current active applications of LLMs for CAD, as well as the domains that are relatively underexplored. Table \ref{tab:industry} presents the industries most commonly mentioned in the reviewed works. As shown in the table, the manufacturing industry currently attracts the most attention of the research community, whereas the shipbuilding industry has garnered relatively little focus. We also include the textile industry in the table which has no notable associated works. Nevertheless, we predict that future works will be associated to this industry due to the relevance of CAD to textile.

\begin{table}[t]
\center
\caption{Number of works attributed to various LLM applications in CAD.}
\renewcommand{\arraystretch}{1}
\begin{tabular}{l|c}
\toprule
\textbf{Application} & \textbf{Number of Works} \\
\midrule
\midrule
Data Generation & 14 \\
CAD Code Generation & 27 \\
Parametric CAD Generation & 20 \\
Image Generation & 1 \\
Model Evaluation & 4 \\
Text Generation & 24 \\
\bottomrule
\end{tabular}
\label{tab:number}
\end{table}

There are also notable limitations faced by the current state-of-the-art techniques. For example, many systems rely on LLMs to interpret the feedback obtained from similarity comparisons between the ground truth and the generated CAD model. In these settings, LLMs can generally determine whether the design is satisfactory, but they often lack the necessary information to understand how it should be improved. To address this gap, some works \cite{alrashedygenerating, DBLP:journals/corr/abs-2406-00144} employed visual question answering pipelines that pose a large number of targeted questions to the model and then synthesize refinement suggestions. Although this strategy provides partial relief, it remains inherently constrained, underscoring the need for more effective and informative feedback mechanisms that can guide LLMs toward actionable geometric corrections in the future.
\section{Future Directions}\label{future}
Leveraging LLMs in CAD domain is becoming increasingly popular. Our survey shows that the research community has already started exploiting the potential of LLMs to address different CAD challenges. However, this area of research is still in its nascency. Due to the practical applications of CAD and growing trend of automation across industries, we expect to see a much wider interest of the research community in this direction in the near future. Whereas we have already pointed towards the interesting explorations for  future research at the intersection of LLMs and CAD throughout the article, we enumerate a few more potential directions in this section based on our literature review. We hope that the discussion below paves the way to more effective and well-directed future research.
\begin{table}[t]
\center
\caption{A summary of the popular datasets used in the literature.}
\renewcommand{\arraystretch}{1}
\begin{tabular}{c|c|c|c|c}
\toprule
\textbf{Dataset} & \textbf{Associated Works} & \textbf{Source Dataset} & \textbf{Type} & \textbf{Public} \\
\midrule
OpenECAD & \cite{YUAN2024104048} & DeepCAD \cite{wu2021deepcad} & Industry & \cmark \\
Omni-CAD & \cite{xu2025cadmllmunifyingmultimodalityconditionedcad} & Onshape & Industry & \cmark \\
Text2CAD 1.0 \& 1.1 & \cite{khan2024text2cad, kolodiazhnyi2025cadrille, chen2025cadreview} & DeepCAD & Industry & \cmark \\
Text-to-CAD & \cite{wang2025text} & DeepCAD, SkexGen \cite{xu2022skexgen} & Industry & \xmark \\
CAD-GPT & \cite{wang2024cad} & DeepCAD & Industry & \xmark \\
BlendNet & \cite{du2024blenderllm} & - & Synthetic & \cmark \\
LLM4CAD & \cite{li2024llm4cad, li2025llm4cad, sun2025large} & - & Synthetic & \xmark \\
CADPrompt & \cite{alrashedygenerating, panta2025meda} & DeepCAD & Industry & \xmark \\
Query2CAD & \citep{DBLP:journals/corr/abs-2406-00144} & - & Synthetic & \cmark \\
CAD-Assistant & \cite{mallis2024cad} & SGPBench \cite{qiu2024can} & Industry & \xmark \\
CAD-Recode & \cite{rukhovich2024cad, kolodiazhnyi2025cadrille} & - & Synthetic & \cmark \\
DeepCAD & \cite{rukhovich2024cad, zhang2025flexcad, zhanggeocad} & - & Industry & \cmark \\
Fusion360 \cite{willis2021fusion} & \cite{rukhovich2024cad, kolodiazhnyi2025cadrille, govindarajan2025cadmium} & - & Industry & \cmark \\
CC3D \cite{Mallis_2023_ICCV} & \cite{rukhovich2024cad, kolodiazhnyi2025cadrille} & - & Industry & \cmark \\
QueryCAD & \cite{kienle2024querycad} & ABC \cite{koch2019abc} & Industry & \xmark \\
SketchGraphs \cite{seff2020sketchgraphs} & \cite{wu2024cadvlm} & - & Industry & \cmark \\
Text2CAD & \cite{li2024cad} & DeepCAD & Industry & \xmark \\
Img2CAD & \cite{you2024img2cad} & ShapeNet \cite{chang2015shapenet} & Synthetic & \xmark \\
ChatCAD & \cite{tang2025chatcad} & - & Synthetic & \xmark \\
ReparamCAD & \cite{10.1145/3610548.3618219} & - & Synthetic & \xmark \\
CADTalk & \cite{Yuan_2024_CVPR, chen2025cadreview} & - & Synthetic \& Industry & \cmark \\
CAD2Program & \cite{CAD2Program} & - & Industry & \xmark \\
CAD-Llama & \cite{li2025cad} & DeepCAD & Synthetic \& Industry & \xmark \\
GenCAD-Code & \cite{doris2025cad} & GenCAD \cite{alam2024gencad} & Synthetic & \cmark \\
Text-to-CadQuery & \cite{xie2025text} & DeepCAD, Text2CAD \cite{khan2024text2cad} & Industry & \cmark \\
CAD-Editor & \cite{yuan2025cad} & DeepCAD & Synthetic & \xmark \\
CCC & \cite{rosnitschek5555076dialogue} & - & Synthetic & \xmark \\
DesignQA & \cite{doris2025designqa} & - & Industry & \xmark \\
MPKG-WT & \cite{hu2025question} & - & Industry & \xmark \\
Seek-CAD & \cite{li2025seek} & ABC & Industry & \xmark \\
CAD2DMD & \cite{valente2025cad2dmd} & - & Synthetic & \xmark \\
DMDBench & \cite{valente2025cad2dmd} & - & Industry & \xmark \\
mrCAD & \cite{mccarthy2025mrcad} & SketchGraphs & Industry & \xmark \\
CADmium & \cite{govindarajan2025cadmium} & DeepCAD & Industry & \cmark \\
DrivAerNet++\cite{elrefaie2024drivaernet++} & \cite{elrefaie2025ai} & - & Industry & \xmark \\
VideoCAD & \cite{manvideocad} & DeepCAD & Synthetic & \xmark \\
CAD-Coder & \cite{guan2025cad} & Text2CAD \cite{khan2024text2cad} & Industry & \xmark \\
CADInstruct & \cite{lv2025cadinstruct} & DeepCAD & Industry & \xmark \\
TriView2CAD & \cite{niu2025creft} & - & Synthetic \& Industry & \xmark \\
ShapeNet\cite{chang2015shapenet} & \cite{yin2025shapegpt} & - & Industry & \xmark \\
CADReview & \cite{chen2025cadreview} & - & Synthetic & \xmark \\
\bottomrule
\end{tabular}
\label{tab:dataset}
\end{table}
\subsection{Application in Interior / Home Design}
A promising future direction for LLMs in CAD is home design, also known as interior design. A few LLM-based tools for home design have already started to emerge \cite{king2023get, ccelen2024design}. For instance, HomeGPT\footnote{\url{https://www.homegpt.app/}} focuses on redesigning individual rooms with AI, making it ideal for completing home makeovers or room-by-room transformations. RoomGPT\footnote{\url{https://www.roomgpt.io/}} is another tool specializing in interior design, similar to ChatGPT, but tailored for creating and visualizing interior design styles and layouts. Users can upload a picture, and these systems generate a \enquote{dream space}. In \cite{ccelen2024design}, I-Design utilized LLMs to transform text input into feasible scene graph designs, considering the relationships between objects. Littlefair et al. \cite{littlefair2025flairgpt} demonstrated that LLMs could be combined with traditional optimization techniques to generate both functional and aesthetically pleasing interior designs. However, some of these tools remain closed-source, and others do not focus on 3D CAD models. Consequently, home design represents a promising, but currently an underexplored area that is expected to attract more research attention in the future.
\subsection{Specific Data Format Generation}
State-of-the-art LLMs are capable of generating a wide range of data formats, including text, image, audio, video, and code. As LLMs continue to evolve, future versions may expand their capabilities to generate even more complex data formats, such as point cloud and even 3D model. Consequently, the scope of LLM applications in data generation extends beyond merely producing text, captions, or textual descriptions. Future research could explore the potential of LLMs to generate these more specialized forms of data, paving the way for advancements in fields like 3D modeling, spatial data generation, and other emerging domains.
\subsection{Building Compliance Checking in AEC}
In the AEC industry, building designs must adhere to a wide range of requirements set forth by codes and standards \cite{ying2024automatic}. Compliance with these regulations is essential to avoid legal issues, delays, and safety hazards \cite{saka2024gpt}. Traditionally, ensuring compliance has been a manual process, requiring labor-intensive checks using 2D drawings and documents, which is time-consuming and costly. Additionally, building codes are complex and frequently updated, further complicating the process.
As a result, the need for Automatic Compliance Checking (ACC) has emerged as a critical solution. ACC involves two main steps: understanding design requirements from textual descriptions and analyzing 3D CAD models or 2D drawings to verify that they meet these requirements. Despite its importance, a universal and scalable ACC system remains a challenge \cite{amor2021promise}. LLMs have already shown some promise in addressing this challenge within the AEC industry. For instance, Zheng and Fischer utilized GPT technologies to enhance natural language-based building information modeling (BIM) searches \cite{zheng2023dynamic}. Additionally, Ying and Sacks \cite{ying2024automatic} explored the use of LLMs for generic building compliance checking using BIM data. Du et al. \cite{du2024text2bim} further leveraged LLMs to convert textual descriptions into code, generating editable BIM models and guiding iterative improvements in model quality. While these efforts focus on building infrastructure, there is a noticeable gap in research that directly applies LLMs to 3D CAD models. However, using LLMs to analyze 3D CAD models or 2D CAD drawings for automatic compliance checking holds significant potential, and further exploration in this area could greatly benefit the AEC industry.
\begin{table}[t]
\center
\caption{Mapping state-of-the-art research to relevant industries in CAD applications.}
\resizebox{\textwidth}{!}{
\begin{tabular}{
>{\centering\arraybackslash}m{1.5cm}|
>{\centering\arraybackslash}m{2cm}|
>{\centering\arraybackslash}m{2cm}|
>{\centering\arraybackslash}m{1.5cm}|
>{\centering\arraybackslash}m{1.2cm}|
>{\centering\arraybackslash}m{0.9cm}|
>{\centering\arraybackslash}m{4cm}|
>{\centering\arraybackslash}m{1.1cm}|
>{\centering\arraybackslash}m{1.5cm}|
>{\centering\arraybackslash}m{1.5cm}|
>{\centering\arraybackslash}m{1.8cm}|
>{\centering\arraybackslash}m{1.2cm}
}
\toprule
\textbf{Industry} & Automotive & Architecture & Shipbuilding & Aerospace & Textile & Manufacturing & Medicine & Electronics & Consumer products & Entertainment & Education \\
\midrule
\textbf{Associated Works} & \cite{YUAN2024104048}, \cite{xu2025cadmllmunifyingmultimodalityconditionedcad}, \cite{du2024blenderllm}, \cite{Makatura2024Large}, \cite{liu20233dall}, \cite{10.1145/3610548.3618219}, \cite{wu2024cadvlm}, \cite{sun2024ai}, \cite{picard2023concept}, \cite{li2024large}, \cite{govindarajan2025cadmium}, \cite{elrefaie2025ai}, \cite{manvideocad}, \cite{lv2025cadinstruct} & \cite{YUAN2024104048}, \cite{xu2025cadmllmunifyingmultimodalityconditionedcad}, \cite{du2024blenderllm}, \cite{makatura2023can}, \cite{Makatura2024Large}, \cite{wu2024cadvlm}, \cite{you2024img2cad}, \cite{zhang2025flexcad}, \cite{sun2024ai}, \cite{picard2023concept}, \cite{li2024large}, \cite{xie2025text}, \cite{govindarajan2025cadmium}, \cite{lv2025cadinstruct}, \cite{madireddy2025large}, \cite{lee5396057automated} & \cite{xu2025cadmllmunifyingmultimodalityconditionedcad} & \cite{YUAN2024104048}, \cite{xu2025cadmllmunifyingmultimodalityconditionedcad}, \cite{du2024blenderllm}, \cite{makatura2023can}, \cite{Makatura2024Large}, \cite{naik2024artificial}, \cite{Yuan_2024_CVPR}, \cite{sun2024ai}, \cite{picard2023concept}, \cite{li2024large}, \cite{govindarajan2025cadmium}, \cite{manvideocad}, \cite{lv2025cadinstruct} & - & \cite{YUAN2024104048}, \cite{xu2025cadmllmunifyingmultimodalityconditionedcad}, \cite{khan2024text2cad}, \cite{wang2025text}, \cite{wang2024cad}, \cite{du2024blenderllm}, \cite{li2024cad}, \cite{li2024llm4cad}, \cite{li2025llm4cad}, \cite{sun2025large}, \cite{alrashedygenerating}, \cite{DBLP:journals/corr/abs-2406-00144}, \cite{makatura2023can}, \cite{Makatura2024Large}, \cite{nelson2023utilizing}, \cite{mallis2024cad}, \cite{rukhovich2024cad}, \cite{jones2025solver}, \cite{naik2024artificial}, \cite{kienle2024querycad}, \cite{ocker2025idea}, \cite{DENG2024221}, \cite{wu2024cadvlm}, \cite{you2024img2cad}, \cite{zhang2025flexcad}, \cite{tang2025chatcad}, \cite{liu20233dall}, \cite{10.1145/3610548.3618219}, \cite{yu2024intelligent}, \cite{khan2024leveraging}, \cite{picard2023concept}, \cite{CAD2Program}, \cite{li2024large}, \cite{li2025cad}, \cite{doris2025cad}, \cite{xie2025text}, \cite{kolodiazhnyi2025cadrille}, \cite{yuan2025cad}, \cite{li2025seek}, \cite{govindarajan2025cadmium}, \cite{guan2025cad}, \cite{lv2025cadinstruct}, \cite{niu2025creft}, \cite{zhanggeocad}, \cite{chen2025cadreview}, \cite{lee5396057automated}, \cite{daareyni2025generative}, \cite{ghosh2025fostering}, \cite{garcia2025llms}, \cite{sadik2025human}, \cite{rosnitschek5555076dialogue}, \cite{panta2025meda} & \cite{makatura2023can}, \cite{Makatura2024Large}, \cite{sun2024ai}, \cite{picard2023concept}, \cite{li2024large} & \cite{makatura2023can}, \cite{Makatura2024Large}, \cite{sun2024ai}, \cite{picard2023concept}, \cite{li2024large}, \cite{valente2025cad2dmd} & \cite{wu2024cadvlm}, \cite{zhang2025flexcad}, \cite{liu20233dall}, \cite{Yuan_2024_CVPR}, \cite{picard2023concept}, \cite{CAD2Program}, \cite{li2024large}, \cite{doris2025cad}, \cite{manvideocad}, \cite{niu2025creft} & \cite{makatura2023can}, \cite{Makatura2024Large}, \cite{govindarajan2025cadmium} & \cite{picard2023concept}, \cite{li2024large}, \cite{10.1115/1.4067318} \\
\bottomrule
\end{tabular}}
\label{tab:industry}
\end{table}
\subsection{Fashion AI and Textile Design}
While much attention has been given to various applications of LLMs, fashion design, particularly in the context of textiles, remains largely unexplored. From Table \ref{tab:industry}, we observe that no state-of-the-art works have focused on leveraging LLMs for textile design. However, CAD plays a crucial role in fashion design, encompassing functions such as creative visualization, technical pattern development, and style information representation \cite{wu2025ai}. CAD in fashion design has streamlined the exchange of information across complex network of communication channels, reducing time, cost, and material usage, while also improving accuracy and garment quality. This is largely due to the digitization of design information, which facilitates communication and decision-making without relying on physical samples. Given these advantages, fashion design presents an exciting opportunity for the application of LLMs in generating 3D CAD models. The potential to combine LLMs with fashion CAD tools could open up new avenues for creativity, efficiency, and innovation in the fashion industry.
\subsection{Open Challenges}
\subsubsection{Multimodal Alignment}
Currently, multimodal alignment across text, images, and parametric or code-based CAD models remains a crucial open challenge. Misalignment among these representations can lead to ambiguous reasoning, incorrect geometric interpretations, and unstable or unpredictable generation quality, ultimately limiting the reliability and controllability of LLM-driven CAD pipelines. In such misalignments, the first major type is image–instruction alignment, where visual content must accurately correspond to the textual descriptions of component positions, proportions, and specified conditions such as symmetry, rotation, and spatial relationships. \cite{kienle2024querycad} leveraged ray casting \cite{glassner1989introduction}, together with additional post-processing steps, to perform image to CAD model alignment. The second major type is script–instruction alignment, where CAD scripts must faithfully implement attributes described in the instructions—including colors, sizes, materials, and other semantic properties that may not be directly observable in the images. Li et al.~\cite{li2024cad} aligned textual descriptions with parametric CAD sequences through a cascading contrastive strategy. In some cases, it is also necessary to align text, image, and point cloud data, in any combination. Hu et al.~\cite{xu2025cadmllmunifyingmultimodalityconditionedcad} proposed an image projection layer for aligning visual data and a point projection layer for aligning point cloud data. In \cite{wang2024cad}, a vision–language projector served as a bridge to align the two modalities.
\subsubsection{Evaluation Inconsistency}
From Section~\ref{sec:eval}, we observe that different works adopt different evaluation metrics, making it difficult to compare methods fairly. Even for commonly used metrics such as Chamfer Distance, these global, non-differentiable measures provide no gradient information and map poorly to local geometric features. As a result, they offer limited guidance for refinement and can hinder the improvement steps performed by LLMs. We propose evaluating models using as many existing metrics as possible, and selecting a subset that best demonstrates model capabilities as a common and universal set of evaluation metrics.
\subsubsection{Data Quality}
From Table~\ref{tab:dataset}, we observe that most existing datasets are derived from DeepCAD, resulting in an overreliance on a single source and potentially constraining the generalizability of current research. Furthermore, a large portion of existing datasets are synthetic, raising concerns about data reliability and the extent to which models trained on such data can generalize to real-world scenarios. In addition, the majority of datasets are private, hindering reproducibility, limiting fair comparison across methods, and slowing down progress in the field. As such, there is a critical need for large-scale, diverse, and publicly accessible real-world datasets that capture a wide range of design settings, modeling workflows, and geometric complexities, thereby enabling more robust evaluation and fostering meaningful advances in LLM-assisted CAD research.
\section{Conclusion}\label{conclusion}
In this comprehensive review, we explored the intersection of LLMs and CAD, highlighting the transformative potential of LLMs in various CAD-related tasks. We began by introducing the significance of CAD in industry, emphasizing its broad applications across different sectors. We then delved into the foundational principles of LLMs in an accessible manner, covering their architecture, training methods, and fine-tuning processes that enable them to perform across diverse domains. We also presented an overview of state-of-the-art LLMs, focusing on both closed-source models like GPT and PaLM and publicly available models such as LLaMA and DeepSeek. Our review underscored the remarkable capabilities of these models in handling complex tasks related to CAD, from data generation to coding and parametric generation, image synthesis, model evaluation and text generation. These advancements open up new possibilities for automating design workflows, improving efficiency, and enhancing creativity in CAD processes. Importantly, we provide a comprehensive analysis of CAD evaluation, reviewing existing metrics and assessment methods in detail. Additionally, we critically analyzed current research trends, pinpointing emerging areas and potential gaps in the existing literature. Finally, we outlined promising future directions, including exploring LLMs in home design, specific data format generation, building compliance checking in AEC, and fashion design and textile applications. These areas offer exciting opportunities for future research and innovation, potentially leading to significant breakthroughs in the integration of LLMs with CAD.
\section{Acknowledgments}
This research was funded (partially) by the Australian Government through the Australian Research Council’s Discovery Early Career Researcher Award, project number DE220101057.

Dr.~Naveed Akhtar is a recipient of the Australian Research Council Discovery Early Career Researcher Award (project number DE230101058) funded by the Australian Government.

\bibliographystyle{ACM-Reference-Format}
\bibliography{main}

\end{document}